\documentclass{article}

\usepackage{microtype}
\usepackage{graphicx}
\usepackage{subfigure}
\usepackage{booktabs} 

\usepackage{amsmath,amsfonts,bm}

\def\eqref#1{equation~\ref{#1}}

\def\1{\bm{1}}

\DeclareMathAlphabet{\mathsfit}{\encodingdefault}{\sfdefault}{m}{sl}
\SetMathAlphabet{\mathsfit}{bold}{\encodingdefault}{\sfdefault}{bx}{n}

\newcommand{\E}{\mathbb{E}}

\newcommand{\R}{\mathbb{R}}

\usepackage{url}
\usepackage{subcaption}
\usepackage{longtable}

\usepackage{hyperref}

\usepackage[accepted]{icml2025}

\usepackage{amsmath}
\usepackage{amssymb}
\usepackage{mathtools}
\usepackage{amsthm}

\usepackage[capitalize,noabbrev]{cleveref}

\theoremstyle{plain}

\theoremstyle{definition}

\theoremstyle{remark}

\usepackage[textsize=tiny]{todonotes}

\icmltitlerunning{Scalable In-Context Learning on Tabular Data via Retrieval-Augmented Large Language Models}

\begin{document}

\twocolumn[
\icmltitle{Scalable In-Context Learning on Tabular Data\\ via Retrieval-Augmented Large Language Models}



\icmlsetsymbol{equal}{*}

\begin{icmlauthorlist}
\icmlauthor{Xumeng Wen}{msra}
\icmlauthor{Shun Zheng}{msra}
\icmlauthor{Zhen Xu}{intern,uc}
\icmlauthor{Yiming Sun}{intern,up}
\icmlauthor{Jiang Bian}{msra}
\end{icmlauthorlist}


\icmlaffiliation{msra}{Microsoft Research Asia, Beijing, China}
\icmlaffiliation{uc}{The University of Chicago, Chicago, IL, USA}
\icmlaffiliation{up}{University of Pittsburgh, Pittsburgh, PA, USA}
\icmlaffiliation{intern}{Zhen and Yiming contributed to this study during their internship at Microsoft Research Asia.}

\icmlcorrespondingauthor{Shun Zheng}{shun.zheng@microsoft.com}

\icmlkeywords{Tabular data, in-context learning, large language models}

\vskip 0.3in
]



\printAffiliationsAndNotice{}  

\begin{abstract}
Recent studies have shown that large language models (LLMs), when customized with post-training on tabular data, can acquire general tabular in-context learning (TabICL) capabilities. These models are able to transfer effectively across diverse data schemas and different task domains.
However, existing LLM-based TabICL approaches are constrained to few-shot scenarios due to the sequence length limitations of LLMs, as tabular instances represented in plain text consume substantial tokens.
To address this limitation and enable scalable TabICL for any data size, we propose retrieval-augmented LLMs tailored to tabular data.
Our approach incorporates a customized retrieval module, combined with retrieval-guided instruction-tuning for LLMs.
This enables LLMs to effectively leverage larger datasets, achieving significantly improved performance across 69 widely recognized datasets and demonstrating promising scaling behavior.
Extensive comparisons with state-of-the-art tabular models reveal that, while LLM-based TabICL still lags behind well-tuned numeric models in overall performance, it uncovers powerful algorithms under limited contexts, enhances ensemble diversity, and excels on specific datasets. 
These unique properties underscore the potential of language as a universal and accessible interface for scalable tabular data learning.
\end{abstract}

\section{Introduction}
\label{sec:intro}

Tabular data, a prevalent data modality comprising rows that represent distinct data instances and columns that define their features and labels, underpins critical machine learning applications across diverse domains, such as healthcare~\citep{johnson2016mimic}, commerce~\citep{bohanec2017PredSales}, and energy~\citep{miller2020EnergyPred}.

A longstanding challenge of learning on tabular data is addressing the heterogeneity in feature and label spaces. Specifically, it is crucial to account for variable-length feature columns and diverse feature types, such as numerical and categorical data with potentially unbounded values. Furthermore, identical feature values may carry entirely different meanings and significance across distinct feature columns and prediction tasks. Lastly, prediction tasks can vary on a case-by-case basis, encompassing both multi-class classification and regression labels.

This heterogeneity challenge has constrained most existing tabular learning approaches to case-by-case optimization or fine-tuning for each new dataset and task. These include early tree-based models~\citep{chen2016XGBoost,ke2017LightGBM,prokhorenkova2018catboost}, later developments in customized neural architectures~\citep{huang2020TabTransformer,arik2021TabNet, katzir2021Net-DNF,gorishniy2021revisit_tab_dnn,gorishniy2024TabR}, and recent explorations in self-supervised learning and pre-training~\citep{yoon2020VIME,somepalli2021SAINT,ucar2021SubTab,bahri2022SCARF,wang2022TransTab,levin2023transfer_tab_nn,zhu2023XTab,yang2023UniTabE,yan2024TP-BERTa,ye2024CM2,ye2024PTaRL}.

Motivated by the emergence of in-context learning (ICL) in large language models (LLMs)~\citep{Brown2020GPT-3}, several studies have begun exploring ICL for tabular data, which we refer to as TabICL in this paper for brevity. Research on TabICL has given rise to two independent threads.
The first pre-trains Transformer~\citep{vaswani2017attention} variants tailored to tabular data using a TabICL objective on synthetic datasets generated from structured causal networks~\citep{hollmann2023TabPFN,hollmann2025TabPFNv2}.
The second post-trains a base LLM using instructions derived from extensive, language-represented real-world tabular datasets, enabling both ICL and zero-shot generalization for new tasks~\citep{wen2024GTL,gardner2024TabuLa}.

These two paradigms each have unique advantages and limitations: the former using numeric representations is highly efficient but confined to the scope of TabICL, while the latter has the potential to seamlessly integrate TabICL with other LLM capabilities, such as conversational interactions, coding, and reasoning, thereby unlocking broader possibilities within a single model.
However, current LLM-based TabICL approaches are limited to few-shot scenarios due to the sequence length constraints of typical LLMs~\citep{wen2024GTL,gardner2024TabuLa}.
Representing a tabular instance in plain text often consumes a substantial number of tokens, significantly restricting their scalability for incorporating more in-context instances.

In this work, we address this limitation by developing a retrieval-augmented generation (RAG)~\citep{lewis2020RAG} approach tailored to tabular data, extending LLM-based TabICL from zero-shot and few-shot scenarios to any-shot settings.
Our approach is built on the assumption that \emph{for a specific test instance, a limited support set can suffice to govern an accurate prediction}, a concept inspired by the core idea of $k$-nearest neighbor algorithms~\citep{fix1951knn,cover1967knn_cls}.
This assumption shifts the optimization focus away from traditional approaches, which aim to compress all training instances into model parameters or hidden states.
Accordingly, we explore the design of a universal, non-parametric retrieval module capable of identifying the most relevant in-context instances across diverse datasets, while also revealing the potential for further enhancing TabICL performance through customized, dataset-specific retrieval policies.
Additionally, we find that aligning LLMs with specific retrieval patterns is crucial, leading to a retrieval-guided generative tabular learning~\citep{wen2024GTL} process for post-training LLMs.

We curate a comprehensive benchmark comprising 29 classification datasets and 40 regression datasets.
These datasets are sourced from different branches of studies~\citep{wen2024GTL,gorishniy2021revisit_tab_dnn,grinsztajn2022tree_gt_tab_nn}, ensuring diversity and fairness by encompassing a wide range of dataset characteristics that may favor different model types, including TabICL models~\citep{wen2024GTL,hollmann2023TabPFN,hollmann2025TabPFNv2}, neural tabular models~\citep{gorishniy2021revisit_tab_dnn,gorishniy2024TabR}, and tree-based models~\citep{chen2016XGBoost,ke2017LightGBM,prokhorenkova2018catboost}.
Our experimental findings reveal the following key takeaways.

\begin{itemize}
\item
The retrieval mechanism significantly enhances LLM-based TabICL by leveraging the scaling of training data size ($D$). Our analysis shows that the median error on held-out datasets ($L$) follows a power-law relationship with $D$: $L(D) = (D_c / D)^{\alpha}$, where $\alpha \sim 0.102$ for classification and $\alpha \sim 0.053$ for regression.
\item
Our approach exhibits remarkable properties, such as uncovering powerful TabICL algorithms comparable to classic models, generating distinct decision boundaries, enhancing ensemble diversity, and achieving superior performance on certain datasets.
\item
Overall, LLM-based TabICL with text-represented data still lags behind well-tuned numeric models. Additionally, no single tabular model consistently outperforms across all evaluation scenarios. While TabPFN, TabR, and LightGBM frequently rank among the top-performing models, they exhibit occasional performance degradations in certain cases, highlighting the importance of developing diverse models in practice.
\item
Through per-dataset analyses and case studies, we find that LLM-based TabICL can be further enhanced by incorporating more effective retrieval policies (e.g., leveraging domain knowledge in practical applications) and by post-training on more diverse feature distributions, similar to the way TabPFN utilizes synthetic datasets. Therefore, we believe this approach, as a special TabICL paradigm, has significant potential to be unlocked.
\end{itemize}

\section{Related Work}
\label{sec:rel_work}

\paragraph{An Overview of Tabular Data Learning}
Tabular data learning has a long-standing history in machine learning research. Prior to the advent of deep learning~\citep{lecun2015DL}, tree-ensemble models~\citep{chen2016XGBoost,ke2017LightGBM,prokhorenkova2018catboost} were developed and quickly became the preferred choice for various tabular data science competitions. Motivated by the success of automatic representation learning in deep neural networks, researchers began exploring effective neural architectures for tabular data~\citep{huang2020TabTransformer,arik2021TabNet,katzir2021Net-DNF,gorishniy2021revisit_tab_dnn}. To this day, some practitioners argue for the superiority of tree-based models, and the competition between tree-based and neural tabular models continues~\citep{shwartzziv2022tab_dl_is_not_all,gorishniy2024TabR}. However, a significant advantage of neural tabular models is their flexibility to integrate with other advanced learning mechanisms, such as self-supervised learning~\citep{yoon2020VIME,somepalli2021SAINT,ucar2021SubTab,bahri2022SCARF} and pre-training~\citep{wang2022TransTab,levin2023transfer_tab_nn,zhu2023XTab,yang2023UniTabE,yan2024TP-BERTa,ye2024CM2}. In recent years, breakthroughs in natural language processing~\citep{kenton2019BERT,Brown2020GPT-3} have led to a trend of combining language models with tabular data learning~\citep{dinh2022LIFT,hegselmann2023TabLLM,yan2024TP-BERTa,ye2024CM2,wen2024GTL,gardner2024TabuLa}.
In this work, we build upon a prior approach~\citep{wen2024GTL} that introduces generative tabular learning that post-trains LLMs to follow TabICL or zero-shot prediction instructions.
Our key contributions include defining a retrieval-augmented TabICL formulation, developing a universal non-parametric retrieval module, and introducing a retrieval-guided training process.

\paragraph{ICL}
ICL emerged from research on scaling language modeling~\citep{kaplan2020ScalingLaws,Brown2020GPT-3}, where it was discovered that LLMs could automatically mimic the behavior of a few examples provided in the context as demonstrations.
Here we summarize some representative follow-up studies that are closely related to this paper, such as selecting effective ICL examples for natural language tasks~\citep{liu2021GoodICLExa,xu2023KNNPrompt}, conducting ICL-style tuning to enhance ICL performance~\citep{wei2022FLAN,gu2023Pre-trainICL}, and understanding the ICL mechanism from the perspectives of Bayesian inference~\citep{muller2021TransDoBayInf,xie2022explainICL} and gradient descent~\citep{dai2023ICLasGD}.
For readers seeking a deeper understanding of ICL, we recommend the comprehensive survey by~\citeauthor{dong2022SurveyICL}.
However, despite the significant advancements of ICL in language modeling, TabICL has received relatively little attention within the tabular data learning community.

\paragraph{TabICL}
As discussed in the introduction, existing TabICL approaches can be broadly categorized into two groups: the TabPFN series~\citep{hollmann2023TabPFN,hollmann2025TabPFNv2} and LLM-based variants~\citep{wen2024GTL,gardner2024TabuLa}.
Both approaches utilize Transformer~\citep{vaswani2017attention} architectures as their backbones but differ in two major aspects.
First, the TabPFN series adopts numerical representations, whereas LLM-based approaches employ text representations. These representations offer distinct strengths and limitations: numerical representations are highly efficient but confined to specific tabular predictions, while text representations can accommodate fewer in-context instances yet integrate seamlessly with other LLM functionalities, infrastructures, and application interfaces.
Second, TabPFN models are trained on synthetic datasets generated from structured causal networks, whereas LLM-based approaches rely on post-training with real-world datasets. In principle, these two data recipes could be combined to achieve complementary benefits. However, to the best of our knowledge, the TabPFN series has not disclosed its synthesized pre-training datasets, particularly the critical data upgrade from TabPFN-v1~\citep{hollmann2023TabPFN} to TabPFN-v2~\citep{hollmann2025TabPFNv2}.
In contrast, LLM-based studies have released fully reproducible pipelines~\citep{wen2024GTL,gardner2024TabuLa}.
In this study, we focus on extending the applicability of LLM-based TabICL from zero-shot and few-shot to any-shot settings and explore the potential of leveraging language as a universal interface for data-driven learning.

\paragraph{RAG}
RAG also emerged from language modeling research to alleviate the limitations of LLMs in handling knowledge-intensive language tasks~\citep{lewis2020RAG}.
This mechanism enables models to leverage external knowledge bases to supplement their representations, producing more accurate and informed responses.
\cite{gao2023SurveyRAG} conducted an extensive survey of numerous follow-up studies on RAG, categorizing them into three main areas: pre-training, fine-tuning, and inference.
In contrast, the application of RAG in tabular data learning remains relatively underexplored.
A recent success in this domain is TabR~\citep{gorishniy2024TabR}, which enhances representations for a neural tabular model by extracting nearest neighbors, thereby improving performance.
In this study, we demonstrate that RAG can enable LLM-based TabICL to effectively handle large-scale datasets.


\section{Formulations of TabICL}
\label{sec:formulation}

We begin by establishing formulations of the TabICL problem, with a focus on the differences in data representation between TabPFN- and LLM-based methodologies.

\paragraph{Notations}
Let $\mathcal{T}$ denote the family of all tabular learning tasks. A typical task $T \sim \mathcal{T}$ is defined as a mapping from an input variable $x^T \in \R^{h_x^T}$ to a label variable $y^T \in \R^{h_y^T}$, expressed as $T: x^T \mapsto y^T$, where $h_x^T$ and $h_y^T$ represent the dimensions of the feature and label spaces, respectively.
We denote the joint distribution of $x^T$ and $y^T$ by $D^T$, such that $(x^T, y^T) \sim D^T$. In practice, a training dataset is constructed by sampling $N$ observations from $D^T$, which we denote as $D^T_{\text{train}} = \{(x^T_i, y^T_i)\}_{i=1}^N$. The training set $D^T_{\text{train}}$ serves as an empirical approximation of the underlying distribution $D^T$.
The goal of traditional supervised learning is to derive an optimized model based on $D^T_{\text{train}}$ that accurately characterizes the conditional distribution $P(y^T_{\text{test}} | x^T_{\text{test}})$ for new samples $(x^T_{\text{test}}, y^T_{\text{test}}) \sim D^T$.


\paragraph{TabICL}
Similar to ICL in language tasks~\citep{Brown2020GPT-3}, TabICL regards the training set $D^T_{train}$ as a set of in-context instances.
Specifically, TabICL aims to characterize the new conditional distribution $P(y^T_{test} | x^T_{test}, D^T_{train})$, namely the label distribution of a testing example conditioned on its own features and other in-context instances, for any task $T$ and empirical observations sampled from $D^T$.
To this end, we need a parameterized model $Q_\theta$ to maximize the following log-likelihood:
\begin{align}
    \E_{ (x^T_{\cdot}, y^T_{\cdot}) \sim D^T, T \sim \mathcal{T} } \left[ \log Q_\theta \left( y^T_{test} | x^T_{test}, D^T_{train} \right) \right].
    \label{eq:TabICL}
\end{align}
In this way, we stimulate $Q_\theta$ to perform TabICL for various tasks.
During the inference stage, $Q_\theta(\cdot | x^{T'}_{test}, D^{T'}_{train})$ directly delivers the label distribution of $y^{T'}_{test}$ for any new task $T' \sim \mathcal{T}$ via a forward pass.

However, it is challenging to design $Q_\theta$ because of the dataset heterogeneity across different tabular learning tasks, namely the feature and label spaces ($\R^{h_x^T} \times \R^{h_y^T}$) differ significantly across different tasks.
To address this challenge, TabPFN models~\citep{hollmann2023TabPFN,hollmann2025TabPFNv2} impose maximal limits on feature and prediction dimensions, constructing separate models for classification and regression tasks.
These design choices ensure broad compatibility with most tabular datasets encountered in practice.

\paragraph{TabICL with LLMs}
In contrast to TabPFN models, which utilize numeric representations of tabular data, recent LLM-based TabICL approaches~\citep{wen2024GTL,gardner2024TabuLa} rely on a serialization function~\citep{dinh2022LIFT,hegselmann2023TabLLM} to convert tabular data into language-based instructions. Consequently, their training process can be framed as maximizing a modified log-likelihood function with an LLM model $Q^{LLM}_{\theta}$:
\begin{align}
    \E_{(x^T_{\cdot}, y^T_{\cdot}), T}
    \left[ \log Q^{LLM}_\theta \left( S_{y} \left( y^T_{test} \right) \middle\vert S_{x} \left( x^T_{test}, D^T_{train} \right) \right) \right],
    \label{eq:TabICL_LLM}
\end{align}
where $S_{y}(\cdot)$ and $S_{x}(\cdot, \cdot)$ are serialization functions that convert tabular data (including heterogeneous numerical and categorical features, as well as variable-sized in-context instances) into sequences of language tokens based on configurable templates. Depending on the chosen configuration, tabular data can be represented in language-like or Markdown-style formats, with the option to exclude $D^T_{\text{train}}$ for enabling zero-shot learning.

\paragraph{Challenges in TabICL with LLMs}
While ongoing efforts aim to scale the context lengths of LLMs~\citep{xiong2023effective,abdin2024phi3}, modern LLMs continue to face significant challenges, including substantial increases in computational overhead~\citep{liu2024RingAttn} and unexpected performance degradation in long-context scenarios~\citep{li2024LongICLBench}.
In the context of TabICL, the challenge of long contexts is particularly pronounced, as tabular instances typically comprise multiple numerical and categorical features, which consume a large number of tokens when represented in a language format. Consequently, the limited context length of $Q^{LLM}_\theta$ severely constrains the number of in-context instances that can be incorporated into $D^T_{\text{train}}$.



\section{TabICL with Retrieval-Augmented LLMs}
\label{sec:method}

To enable scalable TabICL on any-shot settings, we develop a decoupled formulation for~\eqref{eq:TabICL_LLM} as
\begin{align}
\begin{split}
    \E_{(x^T_{\cdot}, y^T_{\cdot}), T}
    &\left[ \log Q^{LLM}_\theta \left( S_{y} \left( y^T_{test} \right) \middle\vert S_{x} \left( x^T_{test}, C_{x^T_{test}} \right) \right) \right], \\
    C_{x^T_{test}} &= \texttt{TabRAG} \left( x^T_{test}, D^T_{train}  \right),
    \label{eq:TabICL_LLM_RAG}
\end{split}
\end{align}
where $C_{x^T_{test}}$ denotes a set of in-context instances selected to support the prediction of $x^T_{test}$, with the size fixed as $N^C$, and we use \texttt{TabRAG} to denote the retrieval module for tabular data that accepts a query example, $x^T_{test}$, and a training set of any scale, $D^T_{train}$, as inputs.

Our formulation decouples essential capabilities for scalable TabICL into two distinct modules, addressing the constraint of limited context length.
The first module, $Q^{LLM}_\theta$, is responsible for predicting the desired outcomes for a query instance $x^T_{\text{test}}$ based on its highly relevant contexts $C_{x^T_{\text{test}}}$.
The second module, \texttt{TabRAG}, focuses on identifying the most relevant and supportive reference instances to facilitate predictions by $Q^{LLM}_\theta$.
To achieve competitive TabICL performance across any new task and any-shot settings, our approach offers opportunities for further improvement in three key areas: 1) the design of \texttt{TabRAG}, 2) TabICL of $Q^{LLM}_\theta$, and 3) the alignment between \texttt{TabRAG} and $Q^{LLM}_\theta$.

\paragraph{The Design of \texttt{TabRAG}}
The primary challenge in designing \texttt{TabRAG} lies in developing a universal retrieval protocol capable of addressing the inherent heterogeneity of tabular data. This includes accommodating variable-sized feature columns, diverse feature types, inconsistent feature importance, and the presence of noise or irrelevant features.
To address these challenges, we propose a non-parametric \texttt{TabRAG} module based on weighted feature-wise similarities as a default retrieval policy, which can be applied to any tabular dataset. Specifically, this requires proper normalization of each feature column and efficient quantification of feature importance, with implementation details provided in Appendix~\ref{app:rag_method}.
In the meanwhile, we observe that a non-parametric, universal retrieval policy may select sub-optimal in-context instances in certain cases, potentially deteriorating TabICL performance.
For instance, as highlighted in case studies of Appendix~\ref{app:case_study}, some slightly adjusted retrieval strategies can significantly boost performance in certain cases.
Therefore, we recommend integrating our basic \texttt{TabRAG} with domain knowledge and data insights in practical applications to further enhance its effectiveness.
We anticipate that ``retrieval engineering'' could emerge as a critical skill for optimizing LLM-based TabICL, akin to the importance of prompt engineering in transforming natural language processing~\citep{liu2023prompt_survey}.

\paragraph{TabICL with $Q^{LLM}_\theta$}
Enhancing the foundational TabICL capabilities remains a key objective in advancing tabular learning with LLMs.
In this work, we upgrade the base LLM used by~\citeauthor{wen2024GTL} from LLaMA-2~\citep{touvron2023llama2} to Phi-3~\citep{abdin2024phi3}, extending the maximum sequence length from 4K to 128K.
Our findings demonstrate the clear benefits of leveraging longer contexts, enabling more effective TabICL performance.
As this study primarily focuses on highlighting the importance and effectiveness of introducing a retrieval mechanism, we provide a few comparisons of different base LLMs in Appendix~\ref{app:exp_base_llm}.
Looking ahead, we anticipate that LLM-based TabICL will continue to benefit from advancements in the LLM field~\citep{liu2024deepseekv3}, further enhancing scalability and unlocking new opportunities for tabular data applications.

\begin{figure*}[t]
\vskip 0.2in
\begin{center}
\centerline{\includegraphics[width=\linewidth]{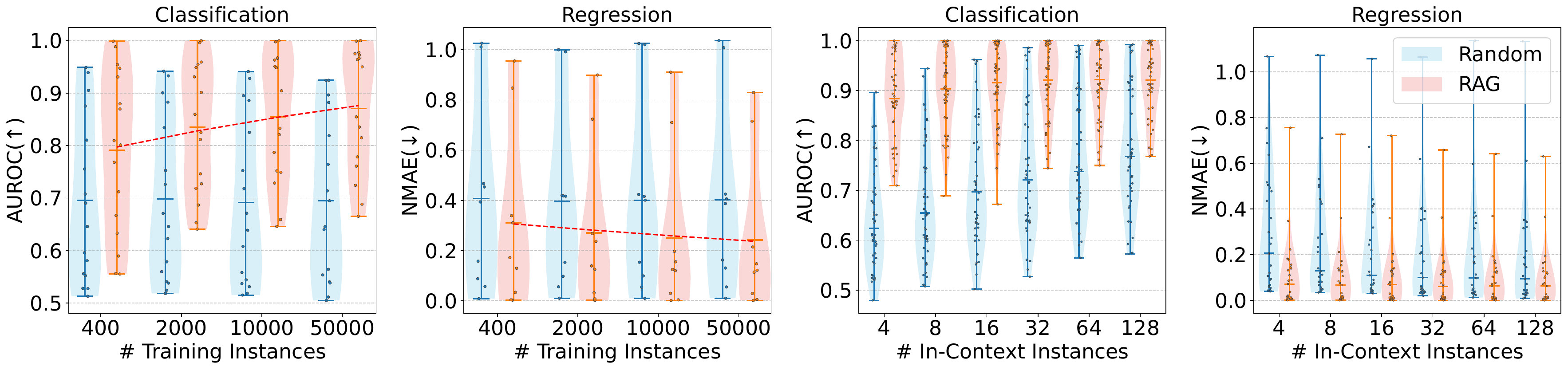}}
\caption{
We investigate the effects of increasing the number of training instances ($|D_{\text{train}}^{T'}|$) and the number of in-context instances per test example ($N^C$) on the TabICL performance of Phi3-GTL models.
In each subplot, we compare the scaling effects of two Phi3-GTL models with different retrieval policies: one that randomly selects in-context instances, denoted as "Random," and the other employing our default \texttt{TabRAG} module, denoted as "RAG".
We use violin plots to visualize the performance distribution across multiple held-out datasets. Additionally, dashed lines are used to emphasize that the median prediction error of our approach follows a power-law relationship with the number of training instances.
}
\label{fig:scaling_pool_ctx}
\end{center}
\vskip -0.2in
\end{figure*}

\paragraph{The Alignment of \texttt{TabRAG} and $Q^{LLM}_\theta$}
To fully unleash the TabICL capabilities of LLMs, it is crucial to align them with the in-context selection patterns of a customized \texttt{TabRAG} module.
Rather than randomly selecting in-context instances, as in~\citep{wen2024GTL}, we post-train LLMs to adhere to the in-context distributions generated by our default \texttt{TabRAG} policy. This alignment results in improved generalization performance on held-out datasets.
These findings emphasize that future post-training of LLM-based TabICL could further benefit from integrating diverse retrieval policies, paving the way for flexible "retrieval engineering" tailored to downstream tabular prediction tasks.

\section{Experiments}
\label{sec:exp}

We conduct extensive experiments to address the following research questions:
1) How effectively can our retrieval mechanism enhance LLM-based TabICL in leveraging large-scale datasets?
2) How does LLM-based TabICL perform in comparison to numeric-based TabICL models and classic tabular models that are well-tuned on a case-by-case basis?
3) What are the unique strengths, current limitations, and potential future directions for LLM-based TabICL?


\subsection{Experimental Setups}
\label{sec:exp_setup}

\paragraph{LLM Post-Training}
We use real-world tabular datasets to post-train a base LLM using generative tabular learning (GTL) objective as did in~\citep{wen2024GTL}.
However, unlike their approach, we adopt Phi-3~\citep{abdin2024phi3} as the base LLM, extending the effective context length from 4K to 128K and aligning it with our default retrieval policy.
Details of this post-training process are provided in Appendix~\ref{app:method_align_rag_llm}. For brevity and clearness, we denote our post-trained model as Phi3-GTL and refer to our approach as RAG+Phi3-GTL throughout the remainder of this paper.

\paragraph{Held-out Datasets}
We compile a comprehensive benchmark from the literature~\citep{gorishniy2021revisit_tab_dnn,grinsztajn2022tree_gt_tab_nn,gorishniy2024TabR,wen2024GTL}, ensuring diverse datasets that may favor different learning paradigms.
To avoid data leakage, we carefully examine and exclude any datasets used during the training of the Phi3-GTL model.
This process results in 29 classification datasets and 40 regression datasets for held-out evaluation, covering a wide range of domains, feature dimensions, types, and distributions.
Details of the data construction process are included in Appendix~\ref{app:data_constr}.

\begin{figure*}[t]
\vskip 0.2in
\begin{center}
\centerline{\includegraphics[width=0.98\linewidth]{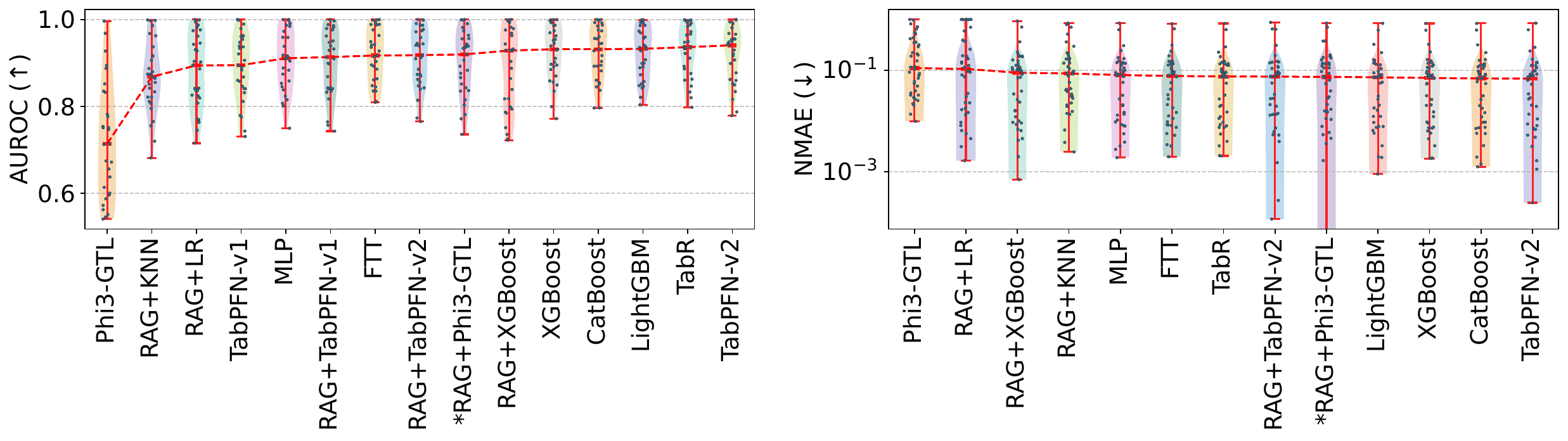}}
\caption{
An overall performance comparison of all models. In the left subplot, we use violin plots to show the AUROC scores of different models across 29 classification tasks, while the right subplot displays the NMAE scores for 40 regression tasks. Models are sorted by their median metric score across the held-out datasets, with dashed lines indicating these median scores in each subplot. Our approach, RAG+Phi3-GTL, is prefixed with a marker (*), for quick identification.
}
\label{fig:overall_comp}
\end{center}
\vskip -0.2in
\end{figure*}

\begin{figure*}[t]
\vskip 0.2in
\begin{center}
\centerline{\includegraphics[width=0.90\linewidth]{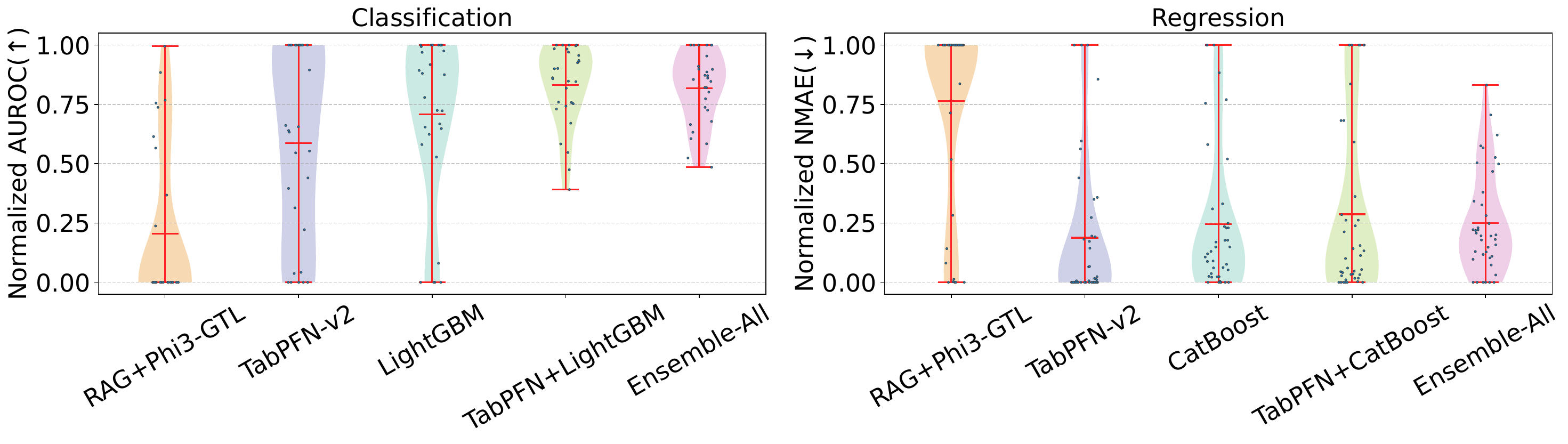}}
\caption{
    Ensemble performance comparisons of RAG+Phi3-GTL, TabPFN-v2, LightGBM, and CatBoost are presented, where normalized AUROC or NMAE scores (min-max normalized across methods for each dataset) are plotted to highlight their relative strengths across multiple datasets, while omitting absolute metric differences.
}
\label{fig:ensemble_res}
\end{center}
\vskip -0.2in
\end{figure*}

\paragraph{Baselines}
We include Phi3-GTL and RAG+KNN as two ablated variants of RAG+Phi3-GTL: the former uses randomly selected in-context instances, while the latter employs the same default retrieval policy but relies on the K-Nearest Neighbors (KNN) algorithm~\citep{fix1951knn,cover1967knn_cls} for prediction.
We compare against TabPFN-v1~\citep{hollmann2023TabPFN}, which supports only classification tasks, and TabPFN-v2~\citep{hollmann2025TabPFNv2}, the state-of-the-art TabICL model utilizing numeric representations.
In addition, our baselines include other representative tabular models such as XGBoost~\citep{chen2016XGBoost}, LightGBM~\citep{ke2017LightGBM}, CatBoost~\citep{prokhorenkova2018catboost}, MLP, FTT~\citep{gorishniy2021revisit_tab_dnn}, and TabR~\citep{gorishniy2024TabR}, all of which are extensively tuned via hyperparameter search for each dataset.
We also include several ``RAG + X'' baselines, where "X" represents models trained and inferred on the selected in-context instances using the same retrieval policy as RAG+Phi3-GTL. These include Logistic Regression (LR), TabPFN-v1, TabPFN-v2, and XGBoost.
These baselines are designed to highlight the TabICL capability of LLMs given limited in-context instances.

\paragraph{Metrics}
For classification tasks, we use the Area Under the Receiver Operating Characteristic curve (AUROC) as the primary evaluation metric. For regression tasks, we employ the Mean Absolute Error normalized by the label mean (NMAE).
Additionally, to compare the relative performance across a group of methods, we utilize group-wise min-max normalized AUROC and NMAE metrics.

\begin{figure*}[t]
\vskip 0.2in
\begin{center}
\centerline{\includegraphics[width=\linewidth]{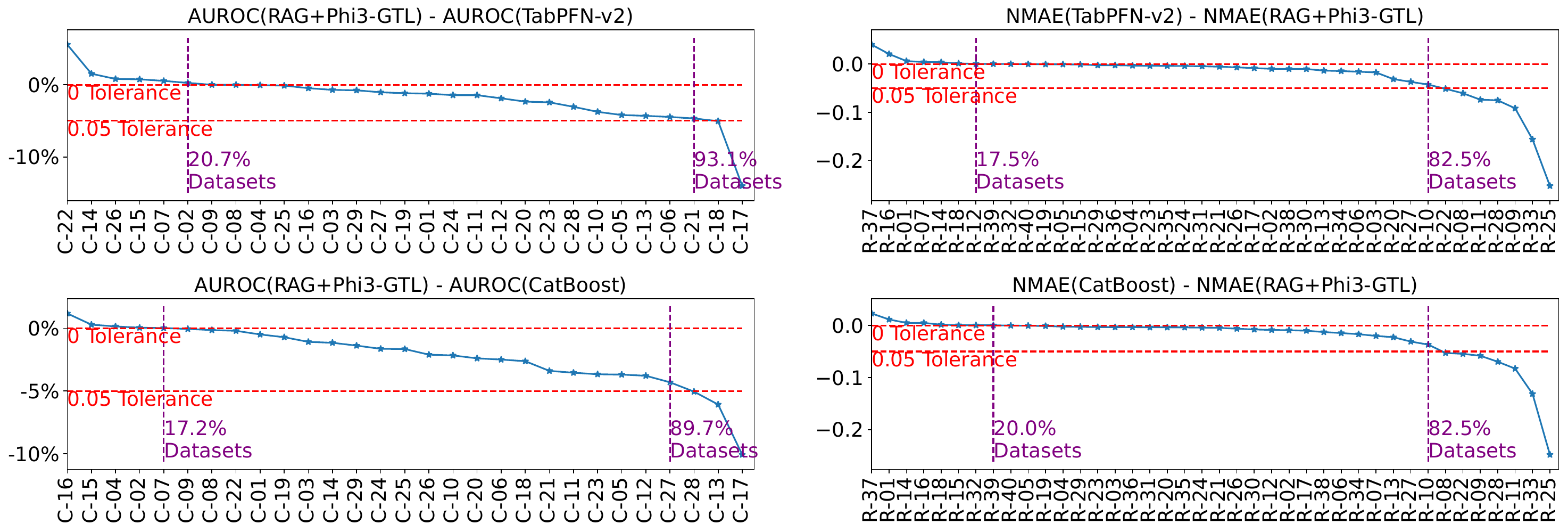}}
\caption{
    Per-dataset performance comparisons between RAG+Phi3-GTL and the two most competitive baselines, TabPFN-v2 and CatBoost, are presented, with dataset IDs sorted by performance gaps. Dashed lines and annotations are used to indicate the proportion of datasets where RAG+Phi3-GTL outperforms these baselines and where it significantly lags behind.
}
\label{fig:per_dataset_comp}
\end{center}
\vskip -0.2in
\end{figure*}

\begin{figure*}[t]
\vskip 0.2in
\begin{center}
\centerline{\includegraphics[width=1.0\linewidth]{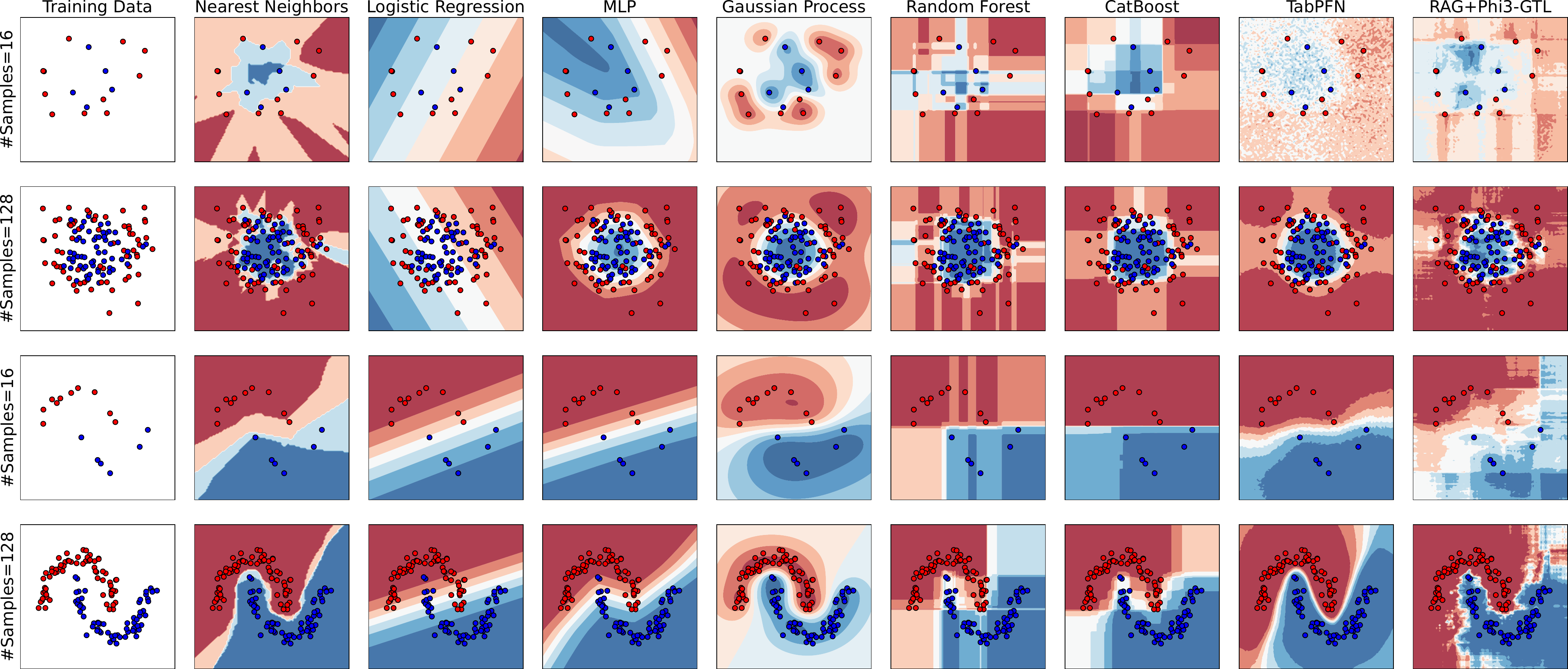}}
\caption{
Decision boundary comparisons of various models, where each row corresponds to a specific set of training instances generated from a given data distribution. The first column visualizes these training instances, while the subsequent columns illustrate the decision boundaries of different models. The top two rows represent the same data distribution but with varying numbers of training instances, whereas the bottom two rows depict a different data distribution.
}
\label{fig:decision_boundary}
\end{center}
\vskip -0.2in
\end{figure*}

\subsection{Scaling with Available Training Instances}
\label{sec:exp_scaling_pool_ctx}

Figure~\ref{fig:scaling_pool_ctx} illustrates the performance variations as the size of the training data and the number of in-context instances increase for our approach under two retrieval policies: Random and RAG (our default retrieval policy).
It is evident that the RAG policy enables Phi3-GTL to effectively leverage larger training datasets, while the Random policy lacks this capability. Specifically, with the RAG policy, the median prediction error demonstrates a power-law relationship with the number of training instances, expressed as $L(D) = (D_c / D)^{\alpha}$. For classification tasks, $L=1-\text{AUROC}$, $D_c \sim 6.05e^{-5}$, and $\alpha \sim 0.102$, whereas for regression tasks, $L=\text{NMAE}$, $D_c \sim 8.05e^{-8}$, and $\alpha \sim 0.053$. This finding highlights a favorable statistical learning characteristic: given a distinguishable feature space and sufficient training instances, the expected prediction error approaches zero.

Moreover, the RAG policy reduces the number of in-context instances required for accurate predictions. As shown in the right two subplots of Figure~\ref{fig:scaling_pool_ctx}, the model using the Random policy benefits significantly from an increased number of in-context instances. In contrast, with the RAG policy, performance often saturates as the number of adaptive in-context instances increases. This indicates that, for most datasets, tens of training instances are sufficient to form a supportive context for inferring the label of a test instance.

\subsection{Overall Comparison}
\label{sec:exp_overall_comp}

Figure~\ref{fig:overall_comp} presents an overall comparison of all models by illustrating the error distributions across held-out datasets. TabPFN-v2 emerges as the most competitive baseline in terms of the median prediction error. However, its wider error bars indicate sub-optimal performance in certain cases. In contrast, well-tuned tree-based models such as LightGBM and CatBoost, as well as neural models like FTT and TabR, demonstrate more robust performance with narrower error distributions.

When comparing our approach, RAG+Phi3-GTL, with these baselines, we observe significant improvements over its ablated variants, Phi3-GTL and RAG+KNN, underscoring the importance of both retrieval and TabICL components. Furthermore, RAG+Phi3-GTL is among the top-performing of ``RAG + X'' baselines, even surpassing RAG+TabPFN-v2 in terms of median prediction performance across held-out datasets.
This highlights the potential of TabICL based on text representations, which can uncover novel and highly effective ICL algorithms by operating in a text space.
Besides, our approach achieves zero NMAE for a specific integer regression task without requiring explicit programming, whereas all numeric models, by default, produce float outputs.
Lastly, RAG+Phi3-GTL still lags behind well-tuned baseline models and TabPFN-v2 in overall performance.

\subsection{Ensemble Results}
\label{sec:exp_ensemble_res}

We further explore the potential of RAG+Phi3-GTL by investigating its contribution to ensemble diversity, as shown in Figure~\ref{fig:ensemble_res}.
When comparing RAG+Phi3-GTL with TabPFN-v2, LightGBM, and CatBoost, we observe that although RAG+Phi3-GTL underperforms the top-performing baselines overall, it exhibits unique strengths in certain scenarios.
Moreover, a comparison of Ensemble-All with TabPFN+LightGBM and TabPFN+CatBoost reveals that these unique strengths translate into ensemble diversity, enhancing the robustness of overall ensemble performance.
These findings highlight the potential of leveraging language as an alternative interface for tabular data learning, complementing existing tabular learning algorithms.

\subsection{Per-dataset Comparisons}
\label{sec:exp_per_ds_comp}

These findings further motivate us to conduct per-dataset comparisons to identify datasets where RAG+Phi3-GTL excels and those where it still underperforms.
Figure~\ref{fig:per_dataset_comp} summarizes these results.
We observe that on approximately 17\%-20\% of datasets, RAG+Phi3-GTL outperforms the state-of-the-art TabICL model, TabPFN-v2, as well as the classic tree-based model, CatBoost, which has been carefully tuned for each dataset.
Furthermore, on over 80\% of datasets, the performance of RAG+Phi3-GTL falls within a small gap of these two competitive baselines.
These results indicate that RAG+Phi3-GTL is already a strong prediction model for most tabular datasets, suggesting that we could not only engage with a well-prepared LLM conversationally but also leverage it to understand tabular data, provide accurate predictions, and offer potential explanations.

\paragraph{Analysis of Failure Cases}
Per-dataset comparison results also prompt an investigation into why the performance of RAG+Phi3-GTL falls short in certain cases. Case studies detailed in Appendix~\ref{app:case_study} provide insights into these failure scenarios.
In summary, most failure cases are attributed to the limitations of the default retrieval policy, which struggles to extract effective in-context instances due to specific data characteristics (e.g., datasets R-25, R-33, and R-27). In these cases, we find that slight adjustments to the retrieval strategy—such as applying alternative numerical normalization methods for feature similarity calculations or leveraging prior knowledge to define instance similarities—can lead to significant performance improvements for RAG+Phi3-GTL.
These findings suggest that a non-parametric, default retrieval mechanism may be insufficient in certain scenarios. Practitioners could potentially achieve better performance by employing "retrieval engineering" (as discussed in Section~\ref{sec:method}).
Additionally, some failure cases, such as dataset C-17, reveal limitations in Phi3-GTL’s ability to perform effective TabICL on specific data distributions, where TabPFN-v2 significantly outperforms. We hypothesize that this gap arises from the limited coverage of data patterns during Phi3-GTL’s post-training phase, which utilized approximately 300 real-world datasets from~\citeauthor{wen2024GTL}. In contrast, TabPFN-v2 likely benefits from pre-training on a much broader family of synthesized datasets.

\subsection{Decision Boundary Analysis}
\label{sec:exp_dec_bound}

Figure~\ref{fig:decision_boundary} compares the decision boundaries of various models across four groups of synthetic instances.
We observe that RAG+Phi3-GTL produces a distinctive, non-smooth decision boundary, which is entirely different from models relying on numeric representations. This boundary reflects case-by-case generalization from known training instances to unseen regions, leaving more uncertain areas when data samples are sparse.

In terms of shape, the decision boundary of RAG+Phi3-GTL bears some resemblance to that of Nearest Neighbors. However, LLM-based TabICL generalizes far beyond a simple rule-based average of neighboring training instances. We hypothesize that this unique behavior arises from the text-based representation of tabular data and the ICL capability of LLMs.
These findings also highlight opportunities for further improving LLM-based TabICL. Specifically, to encourage smoother decision boundaries when sufficient training data is available, one approach could involve generating large-scale synthetic data and fine-tuning LLMs to emulate such behaviors.

\section{Conclusion}
\label{sec:conclu}

In this study, we propose a retrieval-augmented approach to extend LLM-based TabICL from zero-shot and few-shot settings to any-shot scenarios.
This approach explores the potential of using text representations for tabular data learning, enables the creation of unique decision boundaries, and achieves highly competitive prediction performance across most tabular datasets.

Despite the unique strengths and promising potentials, we also acknowledge the limitations of this approach at the current stage, such as the absence of a universally effective retrieval policy, challenges in handling certain long-tail data distributions, and sub-optimal performance in several scenarios.
Given the demonstrated strengths of this approach, we believe that the potential of LLM-based TabICL is still in its early stages, and these limitations present valuable opportunities for future research and development.

\section*{Impact Statement}

This study has the potential to make significant contributions to multiple research communities and practical domains.

For the tabular learning community, this study introduces a novel paradigm for performing TabICL by leveraging text representations of tabular data. This paradigm demonstrates the capability to process tabular datasets across diverse domains and scales, from zero-shot and few-shot to any-shot scenarios. By moving beyond numeric-only representations, our approach opens new avenues for integrating tabular learning into broader, more flexible frameworks, making it possible to unify methodologies across heterogeneous datasets. Furthermore, our retrieval-augmented mechanism provides an adaptable framework for balancing performance and scalability, which may inspire future innovations in retrieval strategies and model architectures.

For the LLM community, this study demonstrates the potential of LLMs to move beyond traditional language-based applications to data understanding and learning tasks involving structured, numeric data. Our findings reveal that LLMs, when augmented with TabICL capabilities, can generalize effectively to tabular datasets and produce competitive results. This highlights the possibility of extending LLMs to domains where structured data plays a central role, such as finance, healthcare, and agriculture. By bridging the gap between text-based and numeric-based data representations, our approach paves the way for a new class of multimodal LLMs capable of unifying text and tabular data learning.

The integration of TabICL capabilities into LLMs may also have ethical and societal implications. On the positive side, these models can democratize access to advanced analytics, enabling users with minimal technical expertise to analyze and understand complex datasets through natural language queries. However, there are potential risks associated with misuse or unintended consequences, such as over-reliance on LLMs for critical decisions, inaccuracies in predictions due to biased training data, and challenges in ensuring transparency and accountability. Researchers and practitioners must prioritize the development of robust evaluation methods, ethical safeguards, and explainability mechanisms to mitigate these risks and ensure responsible deployment.

\bibliography{main}
\bibliographystyle{icml2025}

\newpage
\appendix
\onecolumn

\section{Methodology Details}

\subsection{Our Default Retrieval Policy for Tabular Data}
\label{app:rag_method}
In this section, we describe our default retrieval policy for selecting in-context instances for a query test case.
Our approach employs a non-parametric retrieval mechanism based on the idea of nearest neighbors.
The core idea is to identify the nearest context instances for a given test sample by computing distances between the test instance and the context sample pool (the training set).
We aggregate importance-weighted feature-wise similarity scores as a instance-level similarity score.
This process involves two key components: (1) calculating single-feature distances and (2) aggregating feature distances into instance distances. Below, we elaborate on each component in detail.

\textbf{Calculating Single Feature Distances}
Tabular data typically consists of two types of features: categorical and numerical. To compute distances between samples, we handle each feature type differently:
\begin{itemize}
    \item \textbf{Categorical Features}: For categorical features, the distance between two samples is defined as 1 if their feature values differ and 0 if they are the same.
    \item \textbf{Numerical Features}: For numerical features, we first apply quantile normalization using the statistics of the context pool. Next, we compute the absolute difference between the normalized feature values of the test sample and each context sample. To ensure consistency, we scale these distances to the range [0, 1] using min-max normalization across the context pool.
\end{itemize}
This approach ensures that the distance for each feature between the test sample and any context sample lies within the range [0, 1], providing a standardized measure of similarity.

\textbf{Aggregating Feature-level Distances into Instance-level Distances}
In tabular data, many features may be uninformative, and aggregating feature distances equally can result in selecting context samples that are not representative of important features. To address this, we introduce a parameter-free feature weighting mechanism.
\begin{itemize}
    \item \textbf{Feature Importance Scoring}: We use Pearson Correlation~\citep{cohen2009pearson} to measure the linear relationship between each feature and the target label in the context pool. To capture non-linear relationships, we fit decision trees to each feature and evaluate its contribution to target prediction in the context pool. This process does not require parameter optimization; instead, it iteratively identifies the best feature quantile based on prediction metrics. Since we fit a tree on each single feature, the computation is highly efficient. For implementation, we leverage the Python library PPS (Predictive Power Score)~\citep{florian_wetschoreck_2020_4091345}.
    \item \textbf{Distance Aggregation}: To aggregate feature distances into a sample distance, we employ a weighted L2-norm, defined as~\eqref{appeq:rag_dist}. 
    \begin{equation}
        D_{sample} = \sqrt{\Sigma_{i=1}^{n}(D_{i}^2 \cdot w_i)},
        \label{appeq:rag_dist}
    \end{equation}
    where \( D_i \) is the distance for the \( i \)-th feature, and \( w_i \) is the feature importance weight.
    We combine the two feature importance scores (Pearson Correlation and PPS) to aggregate single-feature distances into sample distances. For instance, if we have a quota of 128 context samples, we allocate half of the quota to the nearest samples based on Pearson Correlation and the other half to the nearest samples based on PPS.
\end{itemize}

\subsection{More Details on Aligning LLMs with Our Default Retrieval Policy}
\label{app:method_align_rag_llm}

To enable generative tabular learning~\citep{wen2024GTL} on data with retrieved contexts, we curated over 300 public datasets from Kaggle, following the collection methodology outlined in their paper. To ensure the integrity of our evaluation, we carefully filtered these datasets to eliminate any potential data contamination between the training and held-out evaluation sets. After filtering, we retained 146 classification datasets and 173 regression datasets. To maximize the utility of these datasets, we expanded each dataset into up to four distinct tasks by designating different columns as task labels. We explored configurations with context sample sizes ranging from 4 to 128, all within a sequence length limit of 16,384 tokens. For each context setting, we randomly selected 16 target samples per task and formatted them using the anonymized template from~\citep{wen2024GTL}. This process yielded a total of over 100,000 data samples for generative tabular learning.

For training, we utilized 16 NVIDIA A100 GPUs. We employed a micro-batch size of 1, a learning rate of 1e-5, and the AdamW optimizer. No warmup or learning rate scheduler was used during training. The entire training process for the Phi-3 Medium base LLM on this dataset took approximately 10 hours.

\section{Dataset Construction}
\label{app:data_constr}

\subsection{Post-Training Datasets for LLMs}
\label{app:data_constr_train}

We initially collected all tabular datasets listed in~\citep{wen2024GTL}. To ensure the integrity of our evaluation, we manually filtered these datasets to remove any potential overlap with benchmark datasets used for evaluation. This careful filtering process resulted in a final collection of 146 classification datasets and 173 regression datasets, which were used exclusively for post-training LLMs.

\subsection{Benchmarking Datasets}
\label{app:data_constr_eval}


To ensure a comprehensive evaluation of our model, we curated datasets from multiple studies focusing on different tabular learning paradigms: LLMs~\citep{wen2024GTL}, neural tabular models~\citep{gorishniy2021revisit_tab_dnn,gorishniy2024TabR}, and tree-based models~\citep{grinsztajn2022tree_gt_tab_nn}.
In total, we prepared 29 classification datasets and 40 regression datasets, enabling a thorough assessment of model performance across diverse tabular data tasks. The detailed dataset information and statics can be found in Table ~\ref{tab:heldout_data}

\textbf{Datasets from GTL~\citep{wen2024GTL}}. GTL is a research initiative focused on post-training LLMs for generalizable inference across diverse tabular tasks. A base LLM is continually trained on a comprehensive collection of 350 datasets and evaluated on an additional 50 datasets, encompassing both classification and regression tasks. This extensive pretraining enables GTL to generalize effectively across various tabular domains, delivering robust performance on a wide range of tasks without requiring task-specific fine-tuning. For our evaluation, we retained 16 test datasets after a strict filtering process based on feature quality and task validity.
We denote the source tag of these datasets as ``LLM''.

\textbf{Datasets from FTT~\citep{gorishniy2021revisit_tab_dnn} and TabR~\citep{gorishniy2024TabR}}. FTT is a widely recognized neural model for tabular learning. TabR represents a significant successor in retrieval-based tabular models. This research proposes an innovative approach that leverages embedding distance-based retrieval methods to enhance the accuracy and efficiency of tabular data processing. The model integrates deep learning architectures with carefully optimized hyperparameters and employs an ensemble strategy, further improving its generalization capabilities across diverse tasks. Through extensive experimentation, TabR has demonstrated superior performance across 11 datasets (those are the same as the ones used in FTT) spanning various scales and complexities, underscoring its adaptability to different tabular data characteristics. For our evaluation, we retained all 11 datasets as heldout datasets, including 5 classification and 6 regression datasets.
We denote the source tag of these datasets as ``Neural''.

\textbf{Datasets from~\citep{grinsztajn2022tree_gt_tab_nn}}, titled Why Do Tree-Based Models Still Outperform Deep Learning on Tabular Data? (referred to as \textbf{Tree}), is a comprehensive research effort that investigates the continued dominance of tree-based over deep learning approaches in tabular data tasks. The study highlights the inherent strengths of tree-based models, which outperform deep learning models across a wide range of tabular tasks, despite significant advancements in neural network architectures. To rigorously evaluate the performance of both model families, the authors curated a diverse collection of small-to-middle-scale datasets, with training samples <= 50,000, categorized into two series: the cat series (containing both numerical and categorical features) and the num series (consisting exclusively of numerical features). 
For our evaluation, we retained the same train-val-test splits as in the original benchmark and used only the first split for consistency. Following the preprocessing approach of TabR, we selected the original dataset when multiple versions of a dataset were available. Additionally, we removed the "Eye Movements" dataset due to potential label leakage concerns and excluded the "isolet" datasets because their large number of features (over 600) would significantly reduce the context quota available for LLMs. After these adjustments, we retained a total of 42 datasets, comprising 15 classification datasets and 27 regression datasets.
We denote the source tag of these datasets as ``Tree''.

\textbf{Post Preprocessing} For all benchmarking datasets, we imposed a restriction on the training size to a maximum of 100,000 samples. For the GTL benchmark datasets, we applied a random split ratio of 8:1:1 for training, validation, and test sets, respectively. For the TabR and Tree benchmarks, we retained the original dataset splits to ensure consistency with prior evaluations. Additionally, to accelerate the evaluation of LLM-based methods, we downsampled each test set randomly to a maximum of 512 samples.
We find this downsampling can ensure statistical significance while improving evaluation efficiency for comparing and investigating different variants of LLMs.

\section{Details of Baselines}



\textbf{Tree-based models} XGBoost, LightGBM, and CatBoost are three widely used Gradient Boosted Decision Tree (GBDT) frameworks, renowned for their ability to optimize speed and performance while handling large datasets, imbalanced data, and missing values. These frameworks are based on the principles of gradient boosting, which builds models sequentially by adding weak learners (decision trees) that correct the errors of prior models. For decades, tree-based models have been the dominant choice and state-of-the-art approach for tabular data tasks.

To ensure a fair comparison, we ran these GBDT models on our benchmarks using hyperparameters tuned for each specific configuration. Following the curated hyperparameter optimization space and search algorithm provided by TabR, we performed hyperparameter optimization using Optuna for 200 trials. This extensive tuning process allowed us to achieve nearly optimal performance for each tabular task. For evaluation, we used the best hyperparameters identified during tuning and ran experiments with 5 different random seeds, reporting the average score across these runs. The hyperparameter tuning space was kept identical to the configuration used in TabR~\citep{gorishniy2024TabR} to maintain consistency and comparability.

\textbf{Neural Baselines}
We evaluate three neural baselines to benchmark performance on tabular data: \textbf{MLP}, \textbf{FT-Transformer}~\citep{gorishniy2021revisit_tab_dnn}, and \textbf{TabR}~\citep{gorishniy2024TabR}. For our experiments, we use the implementations provided by~\citep{gorishniy2024TabR}. We adhere to their hyperparameter tuning process, employing Optuna to conduct 100 trials per dataset within the same tuning space as described in~\citep{gorishniy2024TabR}.

\begin{itemize}
    \item \textbf{MLP} is a simple multi-layer feedforward neural network. We include this baseline to establish a foundational understanding of how deep learning models perform, even at their most basic level.

    \item \textbf{FT-Transformer} (FTT) is a neural architecture specifically designed for tabular data, adapting the transformer framework to handle both categorical and numerical features. Its key innovation lies in its tokenization and processing of tabular data using transformer components traditionally employed in NLP. Specifically, FT-Transformer converts each feature (whether categorical or numerical) into a feature token, treating each feature in a row as an independent input token. This approach enables the model to effectively capture complex relationships within tabular data.

    \item \textbf{TabR} is a retrieval-based neural model for tabular data that integrates in-context retrieval. It encodes both the query dataset and stored datasets into a shared latent space using an embedding function, capturing the structure and relationships between features (both numerical and categorical). The encoded query is compared with pre-stored datasets to identify the most relevant matches in the latent space. By retrieving and leveraging relevant historical data, TabR can make informed predictions on new, unseen datasets without requiring extensive task-specific training. This approach allows TabR to generalize across diverse tabular tasks by utilizing the richness of previously encountered datasets.
\end{itemize}

\textbf{TabICL}
Tabular In-Context Learning (TabICL) methods enable models to adapt to new tabular datasets without explicit retraining by leveraging patterns learned from prior data. We explore two prominent TabICL approaches: \textbf{TabPFN} and \textbf{GTL}.

\begin{itemize}
    \item \textbf{TabPFN} is a prior-data fitted network designed for in-context learning on tabular data. Pretrained on a large corpus of synthetic data, TabPFN learns to recognize common tabular data patterns, such as feature interactions, correlations, and handling of missing values. When presented with a new dataset, TabPFN generalizes in-context without requiring retraining. The initial version, \textbf{TabPFN-v1}~\citep{hollmann2023TabPFN}, supports classification tasks but is limited to datasets with fewer than 1,000 context samples. The recently released \textbf{TabPFN-v2}~\citep{hollmann2025TabPFNv2} extends these capabilities, supporting both classification and regression tasks and handling up to 10,000 context samples through improved prior-data generation, model architecture, and code framework. In our experiments, we found that both TabPFN-v1 and TabPFN-v2 perform effectively with up to 10,000 context samples, with TabPFN-v1 showing improved performance at this scale compared to 1,000 samples. For datasets exceeding 10,000 samples, we randomly select 10,000 samples as the context for both versions.

    \item \textbf{GTL} (Generative Tabular Learning) is a post-training method designed to enhance the capabilities of large language models (LLMs) on tabular datasets. Unlike TabPFN, which relies on embedding representations to understand tabular data, GTL leverages text token representations, harnessing the power of LLMs to interpret and process tabular data. This approach significantly improves zero-shot and in-context learning performance for base LLMs. As demonstrated in~\citep{wen2024GTL}, GTL-enhanced LLMs achieve strong few-shot performance on both classification and regression tasks, though they face challenges due to sequence length limitations inherent to LLMs. In our experiments, we apply GTL to several LLMs using 319 public datasets, carefully filtered from~\citep{wen2024GTL} to ensure no overlap between the post-training datasets and our expanded held-out evaluation datasets. This ensures the integrity of our evaluation and prevents data contamination.
\end{itemize}

\textbf{RAG+KNN}
KNN (K-Nearest Neighbors) is a non-parametric, proof-of-concept model that makes inferences based solely on provided neighbors. In our experiments, we use the retrieved context samples as the neighbors for KNN. By combining KNN with our retrieval mechanism, we can analyze how LLMs learn and reason from context samples compared to simple average predictions.

\textbf{RAG+Other Models}
We evaluate several models enhanced with our retrieval context samples as baselines, including \textbf{RAG+LR} (Logistic Regression and Linear Regression), \textbf{RAG+XGBoost}, and \textbf{RAG+TabPFN}. For each test sample, we fit an individual baseline model using the same retrieval context samples provided to the LLMs. This setup allows us to compare how LLMs perform and make decisions relative to other models when given identical context samples. For \textbf{LR}, we normalize numerical features and labels and apply one-hot encoding to categorical features. For \textbf{XGBoost} and \textbf{TabPFN}, we fit the models directly on the original data. Notably, we use default configurations for LR and XGBoost without hyperparameter tuning to maintain simplicity and focus on the impact of retrieval-augmented context.
In few-shot regression settings, we observe that \textbf{TabPFN-v2} may encounter errors due to its internal normalization when all context labels or numerical feature values are identical. In such cases, we replace the sample prediction with the average of the context labels.

\section{Additional Experimental Results}
\label{app:exp}

\subsection{Retrieval Ablation Analysis: The Importance of Selecting Proper In-Context Instances}
\label{app:exp_rag_abla_test}

\begin{figure}[t]
    \centering
    \includegraphics[width=0.9\columnwidth]{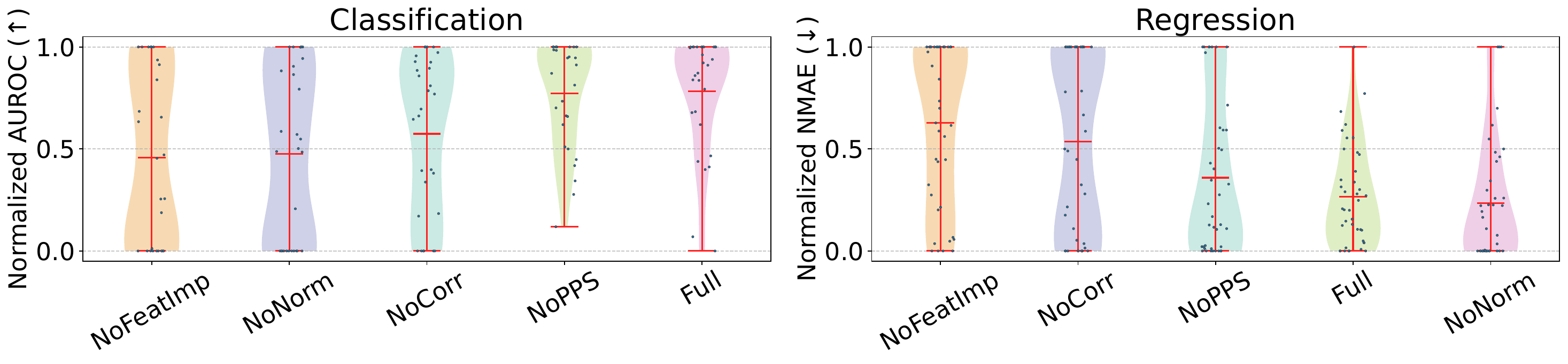}
    \caption{Ablation study of the retrieval mechanism across 29 classification and 40 regression datasets. Normalized metrics compare variants where individual components of the full method (Section~\ref{app:rag_method}) are removed.}
    \label{fig:rag_ablation}
\end{figure}

In this section, we conduct an ablation study of our retrieval mechanism, systematically removing components from the full method (Section~\ref{app:rag_method}) to assess their impact. We evaluate four variants:
\begin{itemize}
\item \textbf{NoFeatImp}: Excludes feature importance scores, aggregating feature distances without feature weighting.
\item \textbf{NoNorm}: Disables numerical feature normalization during distance computation.
\item \textbf{NoCorr}: Uses only Predictive Power Score (PPS) by removing Pearson Correlation for feature importance.
\item \textbf{NoPPS}: Uses only Pearson Correlation by removing PPS for feature importance.
\end{itemize}

Three key findings emerge:
\textbf{Component Necessity}: Removing any component degrades performance, with \textbf{NoFeatImp} exhibiting the largest decline across tasks. This underscores the critical role of feature importance weighting in prioritizing informative features during context selection.

\textbf{Complementary Relationship Measures}: Both linear (Pearson) and non-linear (PPS) metrics contribute uniquely, as performance drops when either is removed (\textbf{NoCorr/NoPPS}).

\textbf{Normalization Effects}: While \textbf{NoNorm} degrades performance in classification tasks, it yields improvements in regression tasks—a counterintuitive divergence in normalization’s impact across tasks (analyzed in Section~\ref{app:case_study}). We hypothesize that normalization’s interaction with feature distributions explains this discrepancy: ill-suited normalization schemes can distort feature distance calculations, particularly in regression scenarios where raw numerical scales may inherently encode meaningful relationships. This observation underscores the necessity of dataset-specific retrieval strategies tailored to the statistical properties of the input data.


\subsection{More Results on Decision Boundary Analysis}
\label{app:exp_dec_bound}
\begin{figure}[t]
    \centering
    \includegraphics[width=0.9\columnwidth]{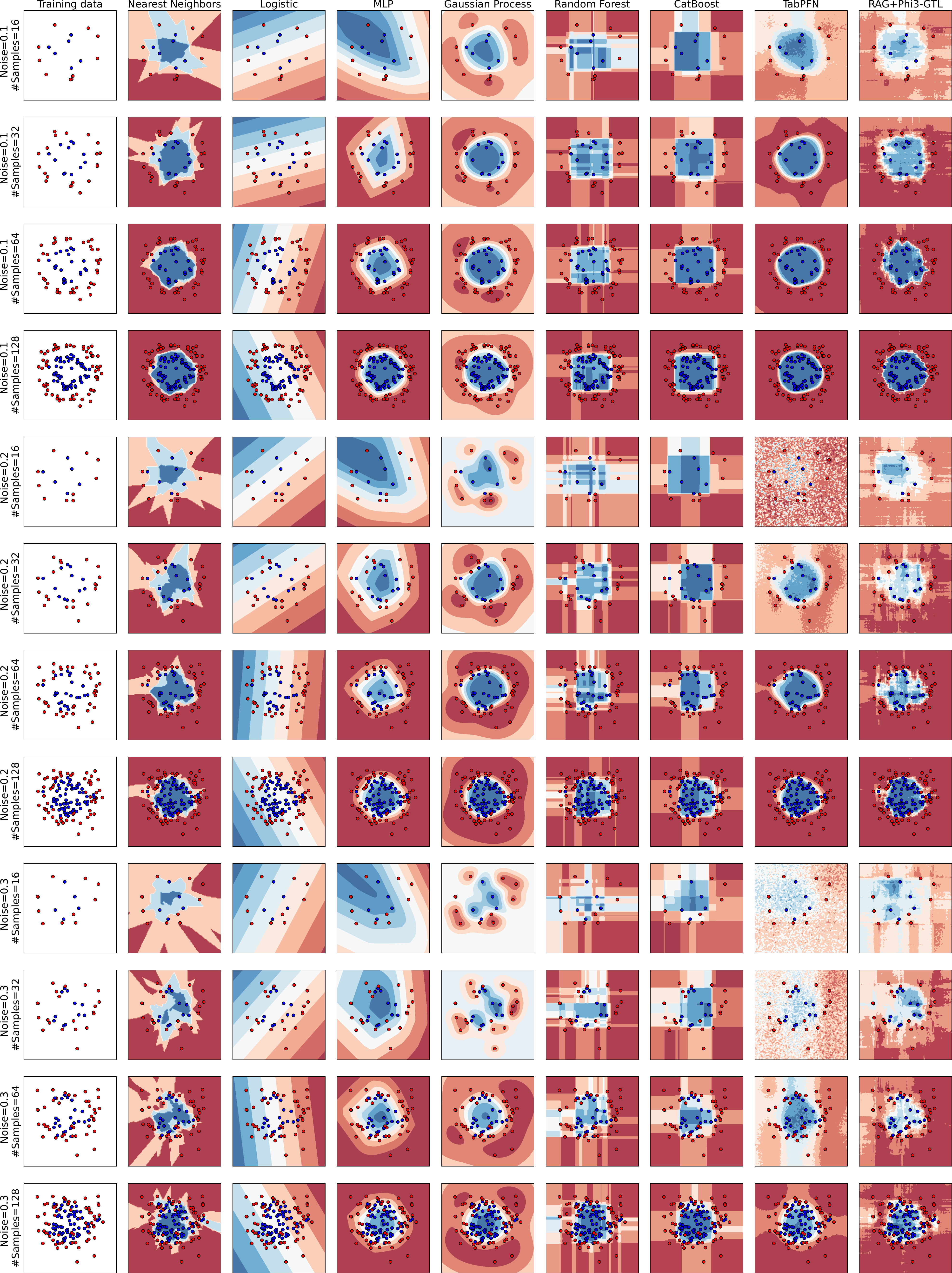}
    \caption{Decision boundaries for the toy dataset \textit{circle}, evaluated across noise levels from 0.1 to 0.3 and training data sizes from 16 to 128.}
    \label{fig:decision_cir}
\end{figure}

\begin{figure}[t]
    \centering
    \includegraphics[width=0.9\columnwidth]{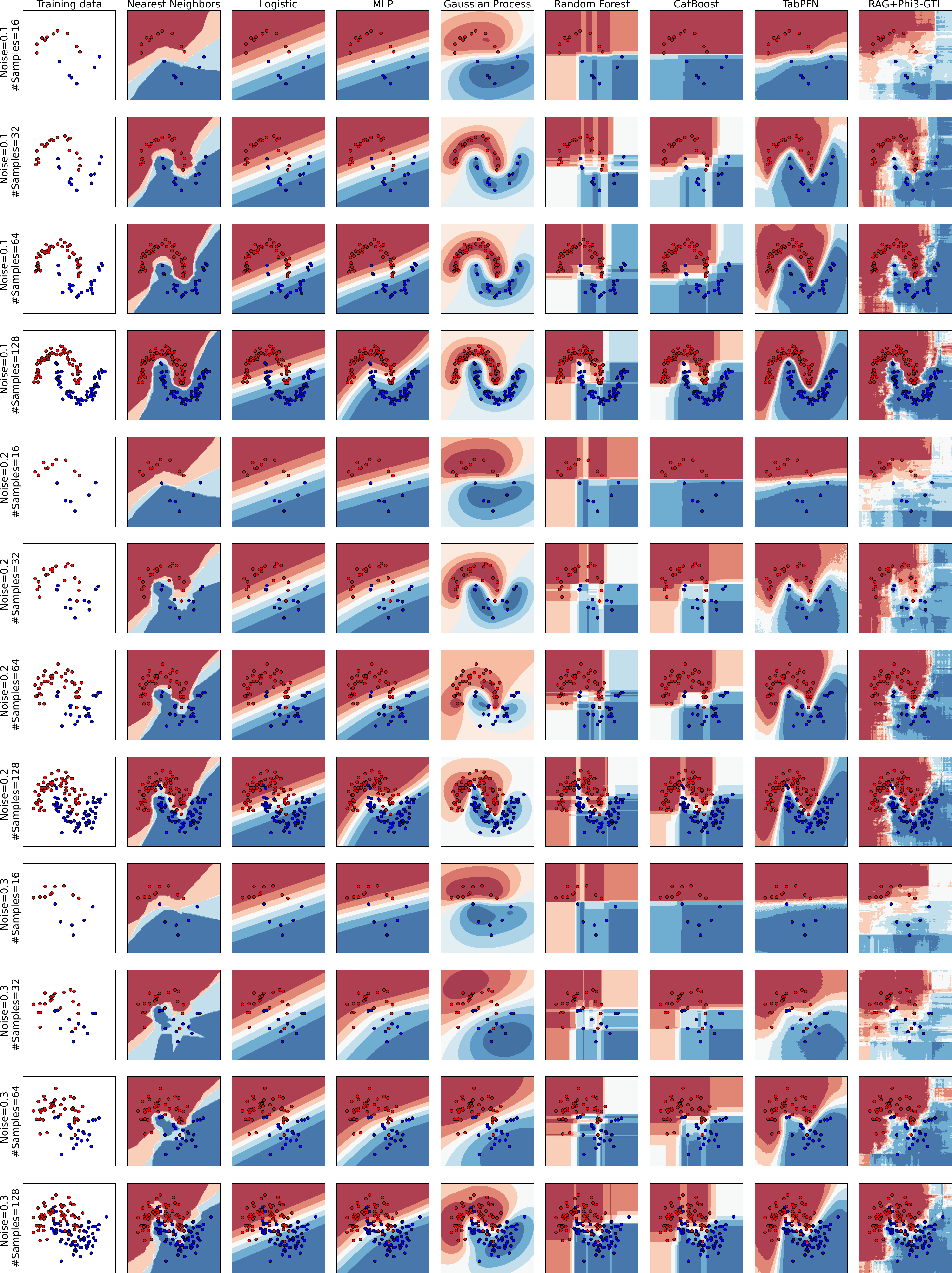}
    \caption{Decision boundaries for the toy dataset \textit{moon}, evaluated across noise levels from 0.1 to 0.3 and training data sizes from 16 to 128.}
    \label{fig:decision_moon}
\end{figure}

\begin{figure}[t]
    \centering
    \includegraphics[width=0.9\columnwidth]{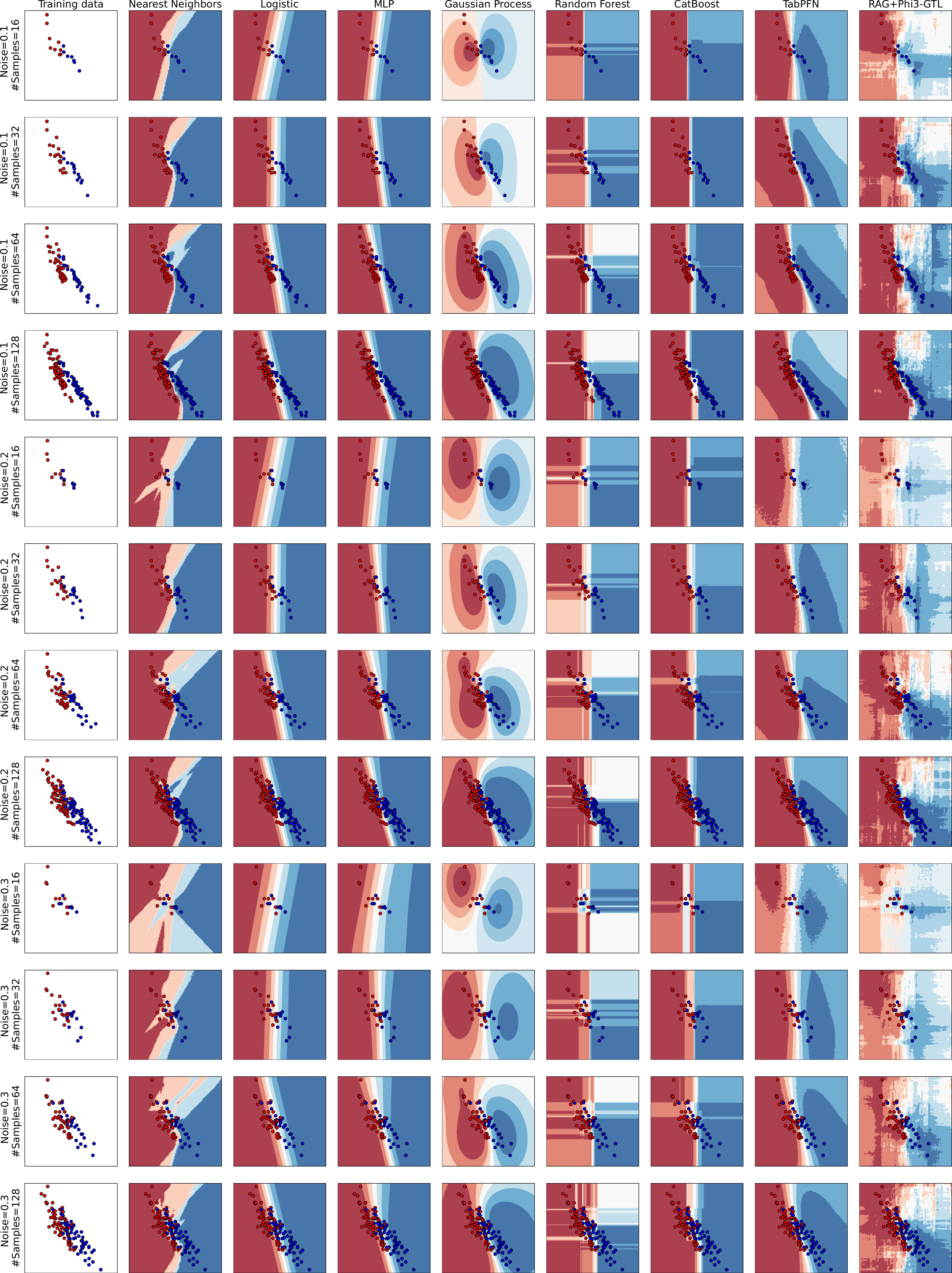}
    \caption{Decision boundaries for the toy dataset \textit{linear\_rotation}, evaluated across noise levels from 0.1 to 0.3 and training data sizes from 16 to 128.}
    \label{fig:decision_line}
\end{figure}

We visualize decision boundaries across diverse data patterns, multiple noise levels, and varying training data sizes in Figure~\ref{fig:decision_cir}~\ref{fig:decision_moon}~\ref{fig:decision_line}. For TabPFN and RAG+Ph3-GTL, the training data serves as the context data samples. Notably, the LLM exhibits unique decision boundaries characterized by higher uncertainty, contrasting with the smoother boundaries of other models. This suggests potential opportunities for LLMs to excel in scenarios with especially limited context or higher noise levels. Interestingly, we also observe a case where TabPFN produces a near-random decision boundary on the circle dataset with noise=0.2 and a training size of 16 (See Figure~\ref{fig:decision_cir}). This indicates that, in specific few-shot scenarios, TabPFN may struggle to fit the data effectively and generalize to test data.

\subsection{More Results using Different Base LLMs}
\label{app:exp_base_llm}

\begin{figure}[t]
    \centering
    \includegraphics[width=0.9\columnwidth]{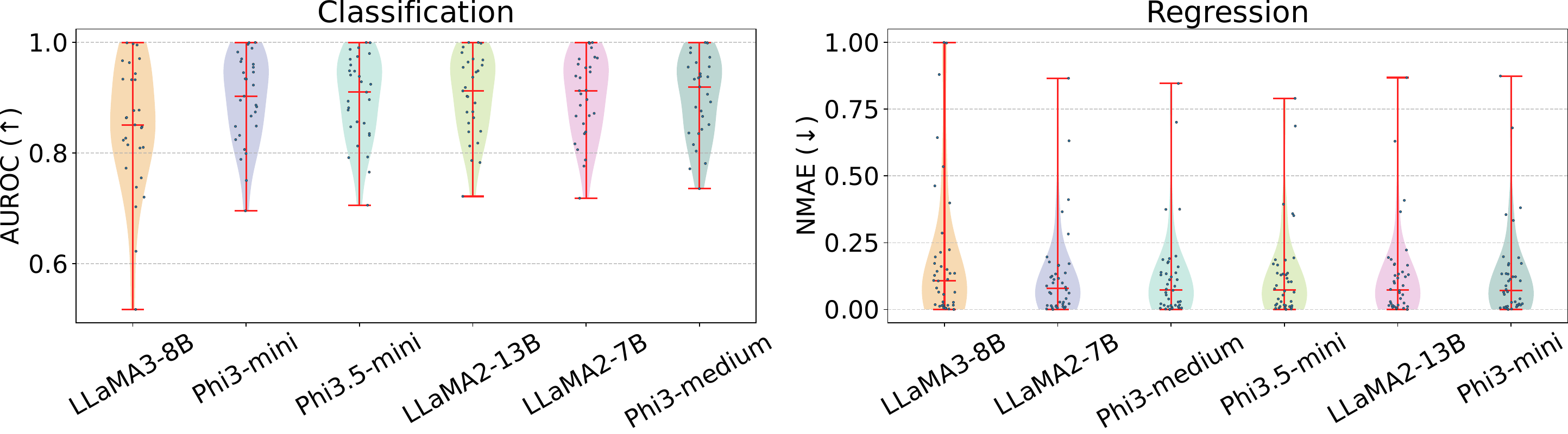}
    \caption{Comparison of base LLM performance with RAG and GTL approaches. Results are evaluated across 29 classification datasets and 40 regression datasets.}
    \label{fig:comp_base_llm}
\end{figure}

We evaluate the retrieval mechanism and GTL across diverse base LLMs, including Phi3-series models (Phi3-mini-128k, Phi3-medium-128k, Phi3.5-mini-128k) and LLaMA-series models (LLaMA2-7B, LLaMA2-13B, LLaMA3-8B). Our results confirm the generalizable effectiveness of these methods across architectures. Two critical findings emerge:

\textbf{Performance Gap in LLaMA3-8B} LLaMA3-8B exhibits a performance gap compared to other base models, producing significantly higher NMAE values in regression tasks (results in Figure~\ref{fig:comp_base_llm} are clipped at 1.0 for visualization). A plausible hypothesis for this discrepancy lies in LLaMA3’s multi-digit tokenization strategy: numerical values are split into three-digit segments (e.g., “123456” → “123” and “456”), whereas LLaMA2 and Phi3 models tokenize each digit individually. This tokenization approach increases the diversity of numerical tokens in the vocabulary, which—combined with standard pretraining protocols—may hinder the model’s ability to learn precise numerical relationships without additional training data. Further investigation is required to isolate the impact of tokenization from other architectural or training factors.

\textbf{Small Models Match Large Counterparts with Retrieval Context} Retrieval-augmented contexts enable smaller models to match the performance of larger ones while maintaining cost efficiency and deployment practicality. This suggests lightweight models, when combined with retrieval mechanisms, offer a viable path to resource-efficient AI deployment without sacrificing accuracy.

\subsection{Comparing with Tabula-8B}
\label{app:exp_comp_tabula}

\begin{figure}[t]
    \centering
    \includegraphics[width=0.9\columnwidth]{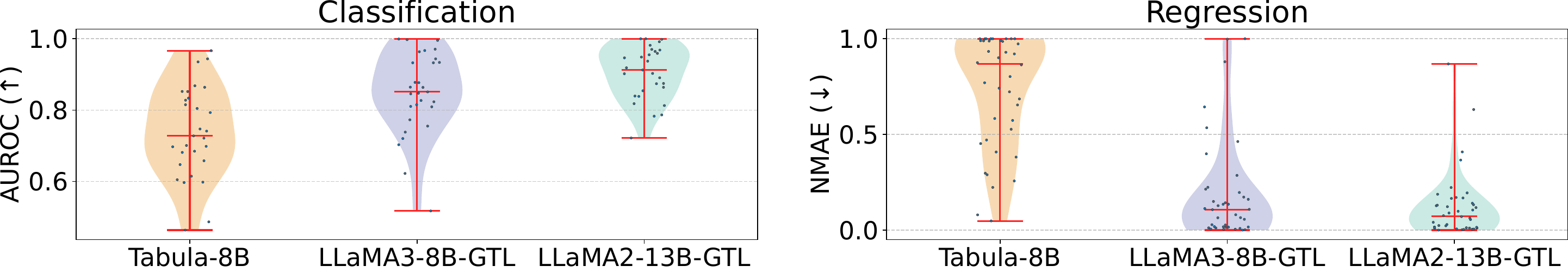}
    \caption{Comparison with Tabula-8B on 29 classification and 40 regression datasets.}
    \label{fig:comp_tabula_ori}
\end{figure}

\textbf{Comparison with Tabula-8B} We compare GTL-enhanced LLMs against Tabula-8B~\citep{gardner2024TabuLa}, a post-trained LLM for tabular data that similarly converts tabular data into language prompts. Tabula-8B builds on LLaMA3-8B but exhibits a significant performance gap relative to LLaMA3-8B-GTL and LLaMA2-13B-GTL in both classification and regression tasks (Figure~\ref{fig:comp_tabula_ori}).
So we mainly use GTL-enhanced LLMs in this work.

\begin{figure}[t]
    \centering
    \includegraphics[width=0.9\columnwidth]{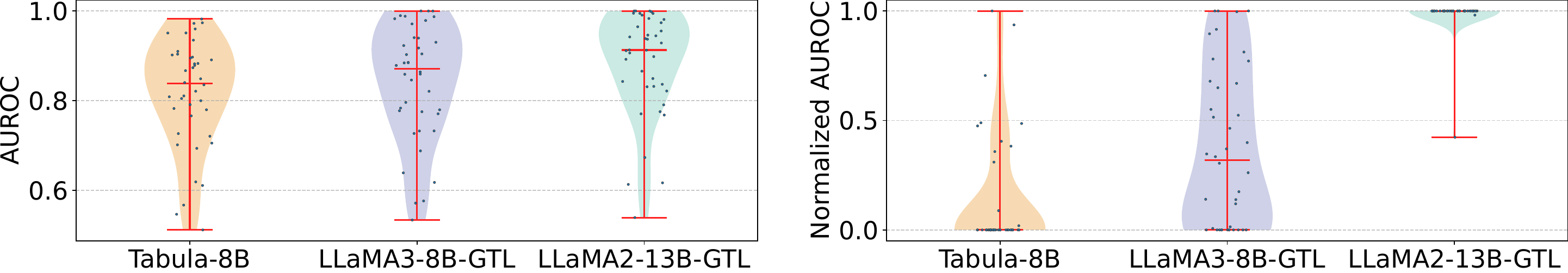}
    \caption{Comparison with Tabula-8B on 40 regression-transformed classification datasets.}
    \label{fig:comp_tabula_reg2cls}
\end{figure}

\textbf{Evaluation on Regression-to-Classification tasks} Moreover, Tabula-8B cannot directly handle regression tasks, instead quantizing numerical labels into four bins (converting regression to classification). To ensure fair comparison, we replicate this methodology by transforming all 40 regression datasets in our benchmark into classification tasks. As shown in Figure~\ref{fig:comp_tabula_reg2cls}, GTL still substantially outperforms Tabula-8B in both absolute and normalized AUROC metrics.

\begin{figure}[t]
    \centering
    \includegraphics[width=0.9\columnwidth]{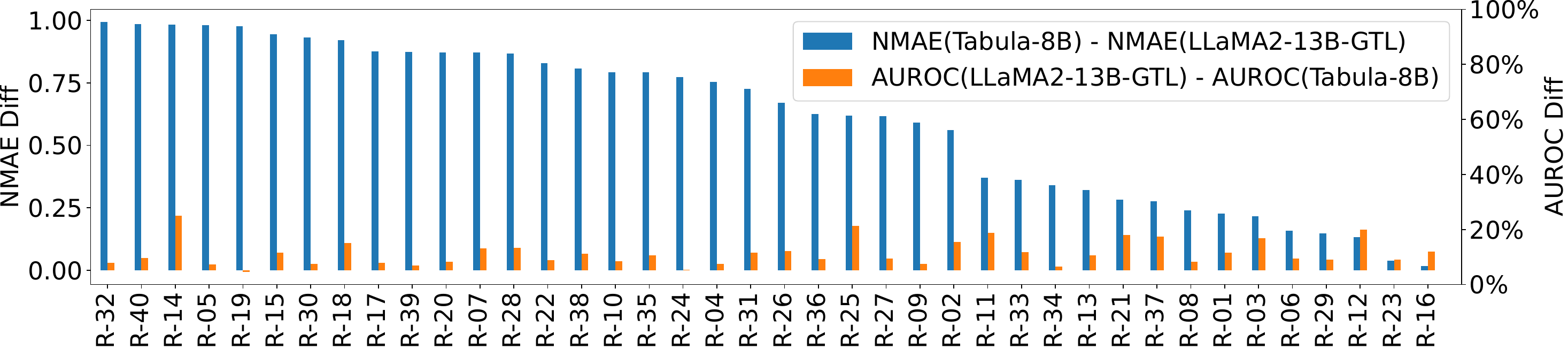}
    \caption{Compare performance gap between Tabula-8B and LLaMA-13B-GTL in regression datasets and the corresponding quantile-transformed classification datasets.}
    \label{fig:comp_reg2cls_per}
\end{figure}

\textbf{Limitations of Regression-to-Classification Conversion} We further investigate whether evaluating on quantized classification datasets reliably reflects performance on original regression tasks. Figure~\ref{fig:comp_reg2cls_per} compares the performance gap between Tabula-8B and LLaMA2-13B-GTL across regression datasets and their quantized classification counterparts. While AUROC gaps remain modest in specific datasets, normalized mean absolute error (NMAE) gaps are often substantial. This underscores the necessity of evaluating models on original regression tasks with true numerical targets—a critical requirement for advancing ensemble methods in tabular learning.

\subsection{Detailed Results of All Models in Overall Comparisons}
\label{app:exp_detail_all_res}
We report detailed results in Table~\ref{tab:detail_res}, including all tuned baselines, TabPFN-v1 and TabPFN-v2, and retrieval-augmented approaches across all 69 benchmark datasets. For tuned baseline, we report the mean metrics over 5 random seeds evaluation after hyperparameters tuning with Optuna. For TabPFN, we conduct in-context learning with full training set (or up to 10,000 samples for larger datasets). For our retrieval methods, we provide as much context as possible (up to 128 context samples) within 16,384 token length limit.

\section{Additional Case Studies}
\label{app:case_study}

\begin{figure}[t]
    \centering
    \includegraphics[width=0.9\columnwidth]{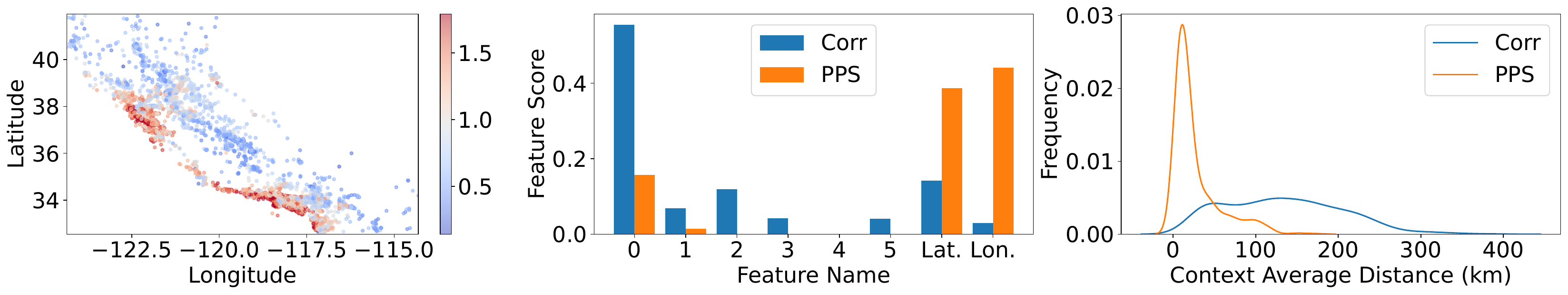}
    \caption{Case Study on the California Dataset. The left figure displays the relationship between features and the label, with label values represented by point color. The middle figure shows feature importance scores computed using linear (Pearson Correlation) and non-linear (PPS) measurements. The right figure presents the distribution of average context distances across all test samples for both linear and non-linear measurements.}
    \label{fig:case_calif}
\end{figure}

\begin{table}
\centering
\caption{Normalized Mean Absolute Error (NMAE) scores for case study datasets. RAG-Tuned denotes results using dataset-specific retrieval methods (see Section~\ref{app:case_study}). The first group highlights datasets where default TabRAG underperformed; customized retrieval contexts significantly improved performance. The second group demonstrates scenarios where LLM in-context learning outperformed both TabPFN and tuned neural models, underscoring its unique strengths for specific data patterns.}
\label{tab:rag_tune}
\begin{tabular}{c|cc|ccc}
\hline
Dataset    & RAG-Tuned+Phi3-GTL & RAG+Phi3-GTL & TabPFN-v2          & CatBoost  & TabR          \\ \hline
R-27       & 0.0951             & 0.1113       & \textbf{0.0742}    & 0.0803    & 0.0775        \\
R-33       & 0.0819             & 0.1904       & \textbf{0.0342}    & 0.0591    & 0.0833        \\
R-25       & 0.1522             & 0.3754       & \textbf{0.1229}    & 0.1273    & 0.1334        \\ \hline
R-14       & -                  & \textbf{0.0045}       & 0.0081             & 0.0094    & 0.0122        \\
R-16       & -                  & \textbf{0.0634}       & 0.0840             & 0.0683    & 0.0729        \\ \hline
\end{tabular}
\end{table}

\textbf{[R-27]} We analyze how feature importance methods influence the selection of context samples on the \textit{regression-num-medium-0-california} dataset, which involves predicting California house prices. As shown in Figure~\ref{fig:case_calif}, the target label exhibits a strong relationship with longitude and latitude. This relationship is not easily captured by linear measurements like Pearson Correlation, whereas the non-linear Predictive Power Score (PPS) assigns high importance scores to both longitude and latitude. As a result, context samples selected using PPS exhibit closer distances to test samples, leading to improved performance. Specifically, compared to default RAG for Phi3-GTL, retrieving nearest context only using PPS can reduce NMAE score from 0.1113 to 0.0951, which is a \textbf{14.6\%} error reduction. This demonstrates that \textbf{both linear and non-linear feature relationships are essential} for identifying important features in tabular data.

\begin{figure}
    \centering
    \includegraphics[width=0.9\columnwidth]{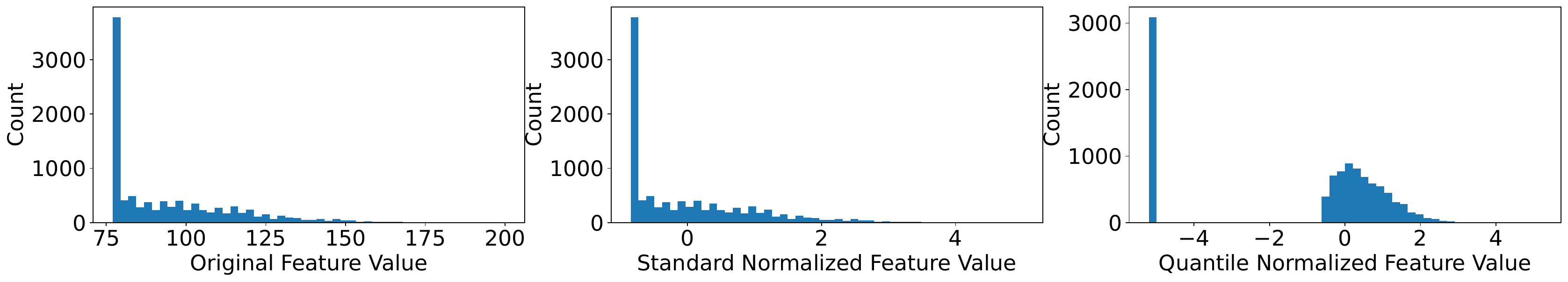}
    \caption{Case Study on the Pol Dataset. The figure displays the distribution of an important feature under three transformations: (left) the original feature values, (middle) standard normalized values, and (right) quantile normalized values}
    \label{fig:case_pol}
\end{figure}

\textbf{[R-33]} The \textit{regression-num-medium-0-pol} dataset describes a telecommunication problem, and we identify one of the most important features using Pearson Correlation between features and the label. As shown in Figure~\ref{fig:case_pol}, the feature distribution varies significantly under different normalization methods. Specifically, the original feature does not follow a true continuous distribution, with over 50\% of its values concentrated at a single point. In this case, quantile normalization disperses this concentrated value and other feature values, which can lead to non-optimal context sample retrieval due to reduced robustness in feature distance calculation. By replacing quantile normalization with standard normalization, the NMAE score dramatically improves from 0.1904 to 0.0819, representing a \textbf{57.0\%} reduction in error. This highlights the significant impact of context quality on LLM performance. It also suggests that users can \textbf{customize retrieval strategies by selecting normalization methods based on feature distributions} and leveraging domain-specific knowledge for specific datasets.

\begin{figure}
    \centering
    \includegraphics[width=0.9\columnwidth]{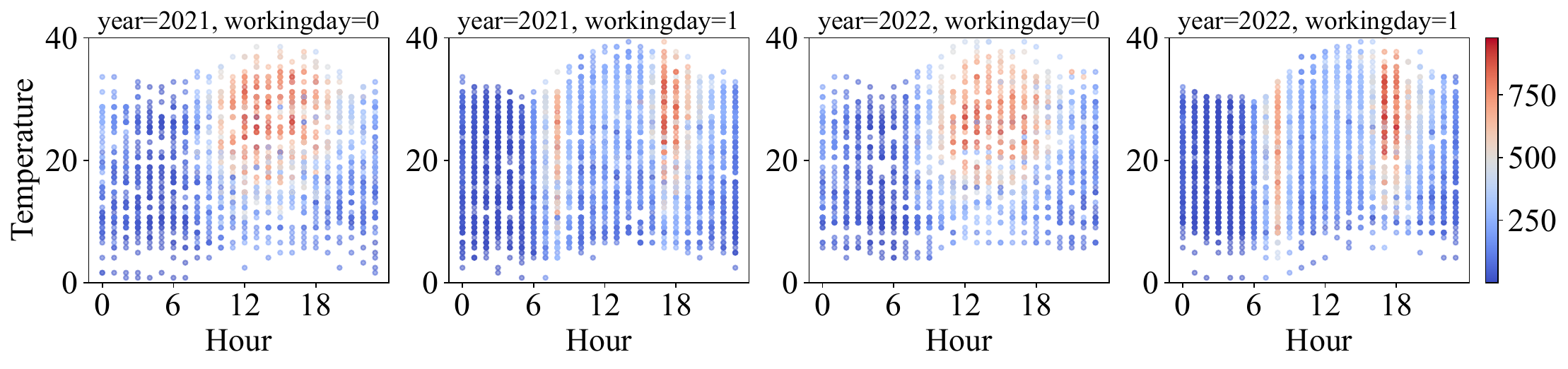}
    \caption{The figure displays the relationship between features (e.g., Temperature and Hour) and the label. It also visualizes bike rental patterns across different years and working-day conditions.}
    \label{fig:case_bike}
\end{figure}

\textbf{[R-25]} We analyze a failure case on the \textit{regression-cat-medium-0-Bike\_Sharing\_Demand} dataset, where the task is to predict the total count of rental bikes. This target is highly correlated with environmental and seasonal factors, including features such as year, season, month, hour, working day, temperature, and humidity. The most important features are hour, temperature, working day, and year. As shown in Figure~\ref{fig:case_bike}, there are distinct patterns between working days and non-working days. On working days, bike rental peaks at 8 am and around 6 pm, while on non-working days, rentals are concentrated between 10 am and 6 pm. Additionally, differences exist between the years 2021 and 2022.

In this case, our retrieval mechanism fails to capture the importance of working day and year because both Pearson Correlation and PPS measure feature importance based on single-feature relationships with the label, rather than feature combinations. We find that restricting context samples to the same year and working-day condition, while retrieving the nearest samples based on other features, reduces the NMAE score from 0.3754 to 0.1522, representing a \textbf{59.5\%} error reduction. This highlights the importance of \textbf{considering feature combinations and leveraging domain knowledge to design customized retrieval strategies} for improved performance.

\begin{table}
\centering
\caption{AUROC scores of Phi3-Medium-GTL and baseline models in RAG scenarios and full data Scenarios on the \textit{classif-cat-medium-0-rl} dataset.}
\label{tab:rl_results}
\begin{tabular}{c|cccc}
\hline
          & Phi3-Medium-GTL & XGBoost & TabPFN-v1 & TabPFN-v2 \\ \hline
RAG       & 0.7811          & 0.7845  & 0.7430    & 0.8621    \\
Full Data & -               & 0.8700  & 0.7307    & 0.9212    \\ \hline
\end{tabular}
\end{table}

\textbf{[C-17]} The \textit{classif-cat-medium-0-rl} dataset is another failure case for our method. As the dataset is anonymized, we lack access to specific feature or task information. However, from the results comparison in Table~\ref{tab:rl_results}, we observe that with retrieved context, Phi-3-Medium-GTL achieves performance comparable to XGBoost and outperforms TabPFN-v1. Interestingly, TabPFN-v2 significantly improves performance on this dataset in both RAG and full data scenarios compared to TabPFN-v1, even surpassing fully tuned XGBoost by over 5\% in AUROC. This improvement may be attributed to the diverse patterns in TabPFN-v2's prior-data. These findings suggest that \textbf{generating synthetic data with diverse feature distributions and feature-label relationships could further enhance the capabilities of LLMs}.

{
\clearpage
\centering
\fontsize{8}{9}\selectfont
\begin{longtable}[t]{c|p{3cm}|p{0.7cm}|c|p{0.6cm}|p{0.5cm}|p{0.5cm}|p{0.5cm}|p{0.5cm}|p{0.7cm}|p{0.7cm}|p{1.1cm}|p{1.2cm}}
\caption{
Data statistics of held-out datasets used in this paper.
}
\label{tab:heldout_data}
\\
\toprule
 Index & Dataset Abbr. & Data Source Tag & Task & \# Train Ex. & \# Test Ex. & \# Class. & \# Num. Feat. & \# Cat. Feat. &
     Max Label Value & Min Label Value &
     Max Feat. Value & Min Feat. Value \\
\midrule
 C-01 & go-to-college-dataset & LLM & Cls. & 720 & 90 & 2 & 4 & 6 & 1 & 0 & 9.41e+06 & 20 \\
 C-02 & datasets-in-it-ops-applied-ai & LLM & Cls. & 720 & 90 & 3 & 3 & 4 & 2 & 0 & 1 & 0 \\
 C-03 & maternal-health-risk-data & LLM & Cls. & 729 & 92 & 3 & 6 & 0 & 2 & 0 & 160 & 6 \\
 C-04 & iris-dataset-extended & LLM & Cls. & 864 & 108 & 3 & 19 & 1 & 2 & 0 & 299.9 & -1.55 \\
 C-05 & telecom-churn-datasets & LLM & Cls. & 2399 & 301 & 2 & 15 & 4 & 1 & 0 & 395 & 0 \\
 C-06 & stroke-prediction-dataset & LLM & Cls. & 3679 & 461 & 2 & 3 & 7 & 1 & 0 & 267.76 & 0.08 \\
 C-07 & predicting-credit-card-customer-attrition-with-m & LLM & Cls. & 7291 & 512 & 2 & 14 & 5 & 1 & 0 & 34516 & 0 \\
 C-08 & factors-affecting-children-anemia-level & LLM & Cls. & 9457 & 512 & 4 & 4 & 11 & 3 & 0 & 218 & 1 \\
 C-09 & stellar-classification-dataset-sdss17 & LLM & Cls. & 72000 & 512 & 3 & 9 & 6 & 2 & 0 & 58932 & -9999 \\
 C-10 & churn & Neural & Cls. & 6400 & 512 & 2 & 7 & 4 & 1 & 0 & 2.51e+05 & 0 \\
 C-11 & adult & Neural & Cls. & 26048 & 512 & 2 & 6 & 8 & 1 & 0 & 1.48e+06 & 0 \\
 C-12 & otto & Neural & Cls. & 39601 & 512 & 9 & 93 & 0 & 8 & 0 & 352 & 0 \\
 C-13 & higgs-small & Neural & Cls. & 62751 & 512 & 2 & 28 & 0 & 1 & 0 & 18.43 & -2.97 \\
 C-14 & covtype & Neural & Cls. & 100000 & 512 & 7 & 10 & 44 & 6 & 0 & 7173 & -166 \\
 C-15 & classif-num-medium-0-wine & Tree & Cls. & 1787 & 512 & 2 & 11 & 0 & 1 & 0 & 440 & 0 \\
 C-16 & classif-num-medium-0-phoneme & Tree & Cls. & 2220 & 512 & 2 & 5 & 0 & 1 & 0 & 3.99 & -3.04 \\
 C-17 & classif-cat-medium-0-rl & Tree & Cls. & 3479 & 512 & 2 & 5 & 7 & 1 & 0 & 1448 & 0 \\
 C-18 & classif-cat-medium-0-KDDCup09\_upselling & Tree & Cls. & 3589 & 512 & 2 & 34 & 15 & 1 & 0 & 1.94e+07 & -6.97e+06 \\
 C-19 & classif-num-medium-0-kdd\_ipums\_la\_97-small & Tree & Cls. & 3631 & 512 & 2 & 20 & 0 & 1 & 0 & 999999 & -9900 \\
 C-20 & classif-num-medium-0-bank-marketing & Tree & Cls. & 7404 & 512 & 2 & 7 & 0 & 1 & 0 & 81204 & -3372 \\
 C-21 & classif-num-medium-0-MagicTelescope & Tree & Cls. & 9363 & 512 & 2 & 10 & 0 & 1 & 0 & 575.24 & -457.92 \\
 C-22 & classif-cat-medium-0-compass & Tree & Cls. & 10000 & 512 & 2 & 8 & 9 & 1 & 0 & 9485 & -578 \\
 C-23 & classif-cat-medium-0-electricity & Tree & Cls. & 10000 & 512 & 2 & 7 & 1 & 1 & 0 & 1 & 0 \\
 C-24 & classif-num-medium-0-credit & Tree & Cls. & 10000 & 512 & 2 & 10 & 0 & 1 & 0 & 250000 & 0 \\
 C-25 & classif-num-large-0-jannis & Tree & Cls. & 40306 & 512 & 2 & 54 & 0 & 1 & 0 & 255 & -132.74 \\
 C-26 & classif-cat-large-0-covertype & Tree & Cls. & 50000 & 512 & 2 & 10 & 44 & 1 & 0 & 7140 & -140 \\
 C-27 & classif-cat-large-0-road-safety & Tree & Cls. & 50000 & 512 & 2 & 29 & 3 & 1 & 0 & 899030 & -6.30 \\
 C-28 & classif-num-large-0-Higgs & Tree & Cls. & 50000 & 512 & 2 & 24 & 0 & 1 & 0 & 20.86 & -2.97 \\
 C-29 & classif-num-large-0-MiniBooNE & Tree & Cls. & 50000 & 512 & 2 & 50 & 0 & 1 & 0 & 1.60e+07 & -12790 \\
 \midrule
 R-01 & alcohol-effects-on-study & LLM & Reg. & 751 & 95 & - & 5 & 17 & 20 & 0 & 75 & 0 \\
 R-02 & airfoil-selfnoise-dataset & LLM & Reg. & 1081 & 136 & - & 5 & 0 & 140.99 & 103.38 & 20000 & 0 \\
 R-03 & abalone-age-prediction & LLM & Reg. & 2556 & 320 & - & 7 & 3 & 29 & 1 & 565.1 & 0 \\
 R-04 & dummy-advertising-and-sales-data & LLM & Reg. & 3287 & 412 & - & 3 & 1 & 364.08 & 31.20 & 100 & 3.13e-05 \\
 R-05 & co2-emission-by-vehicles & LLM & Reg. & 5316 & 512 & - & 6 & 5 & 522 & 96 & 69 & 0.9 \\
 R-06 & melody-metrics-decoding-song-popularity & LLM & Reg. & 5774 & 512 & - & 5 & 1 & 100 & 0 & 2022 & 1.67e-05 \\
 R-07 & car-prices-poland & LLM & Reg. & 84907 & 512 & - & 3 & 6 & 2.4e+06 & 900 & 2.50e+06 & 0 \\
 R-08 & california & Neural & Reg. & 13209 & 512 & - & 8 & 0 & 5.00 & 0.15 & 35682 & -124.35 \\
 R-09 & house & Neural & Reg. & 14581 & 512 & - & 16 & 0 & 5e+06 & 0 & 7.32e+06 & 0 \\
 R-10 & diamond & Neural & Reg. & 34521 & 512 & - & 6 & 3 & 3.73 & -0.90 & 79 & 0 \\
 R-11 & black-friday & Neural & Reg. & 100000 & 512 & - & 4 & 5 & 2.42 & -2.26 & 20 & 0 \\
 R-12 & microsoft & Neural & Reg. & 100000 & 512 & - & 131 & 5 & 4 & 0 & 9.00e+08 & -2.08e+09 \\
 R-13 & weather-small & Neural & Reg. & 100000 & 512 & - & 118 & 1 & 64 & -39 & 1.06e+05 & -9999 \\
 R-14 & regression-cat-medium-0-analcatdata\_supreme & Tree & Reg. & 2836 & 512 & - & 2 & 5 & 2.30 & 0 & 1988 & 0 \\
 R-15 & regression-cat-medium-0-Mercedes\_Benz\_Greener \_Manufacturing & Tree & Reg. & 2946 & 512 & - & 0 & 359 & 265.32 & 72.11 & - & - \\
 R-16 & regression-num-medium-0-wine\_quality & Tree & Reg. & 4547 & 512 & - & 11 & 0 & 9 & 3 & 366.5 & 0 \\
 R-17 & regression-num-medium-0-cpu\_act & Tree & Reg. & 5734 & 512 & - & 21 & 0 & 99 & 0 & 2.53e+06 & 0 \\
 R-18 & regression-cat-medium-0-visualizing\_soil & Tree & Reg. & 6048 & 512 & - & 3 & 1 & 40 & 1 & 166.01 & -0.01 \\
 R-19 & regression-cat-medium-0-yprop\_4\_1 & Tree & Reg. & 6219 & 512 & - & 42 & 20 & 1 & 0 & 1 & 0 \\
 R-20 & regression-num-medium-0-sulfur & Tree & Reg. & 7056 & 512 & - & 6 & 0 & 0.95 & 0 & 1 & 0 \\
 R-21 & regression-cat-medium-0-Brazilian\_houses & Tree & Reg. & 7484 & 512 & - & 8 & 3 & 13.93 & 6.21 & 1.12e+06 & 0 \\
 R-22 & regression-num-medium-0-Ailerons & Tree & Reg. & 9625 & 512 & - & 33 & 0 & 0 & -0.00 & 977 & -872 \\
 R-23 & regression-num-medium-0-MiamiHousing2016 & Tree & Reg. & 9752 & 512 & - & 13 & 0 & 14.79 & 11.18 & 1.60e+05 & -80.54 \\
 R-24 & regression-cat-medium-0-OnlineNewsPopularity & Tree & Reg. & 9999 & 512 & - & 45 & 14 & 12.65 & 1.61 & 843300 & -1 \\
 R-25 & regression-cat-medium-0-Bike\_Sharing\_Demand & Tree & Reg. & 10000 & 512 & - & 6 & 5 & 977 & 1 & 57.00 & 0 \\
 R-26 & regression-cat-medium-0-house\_sales & Tree & Reg. & 10000 & 512 & - & 15 & 2 & 15.49 & 11.23 & 1.65e+06 & -122.51 \\
 R-27 & regression-num-medium-0-california & Tree & Reg. & 10000 & 512 & - & 8 & 0 & 1.79 & 0.14 & 35682 & -124.35 \\
 R-28 & regression-num-medium-0-elevators & Tree & Reg. & 10000 & 512 & - & 16 & 0 & 0.08 & 0.01 & 973 & -966 \\
 R-29 & regression-num-medium-0-fifa & Tree & Reg. & 10000 & 512 & - & 5 & 0 & 12.77 & 6.22 & 1.63e+18 & 16 \\
 R-30 & regression-num-medium-0-house\_16H & Tree & Reg. & 10000 & 512 & - & 16 & 0 & 13.12 & 0 & 3.49e+06 & 0 \\
 R-31 & regression-num-medium-0-houses & Tree & Reg. & 10000 & 512 & - & 8 & 0 & 13.12 & 9.62 & 35682 & -124.35 \\
 R-32 & regression-num-medium-0-medical\_charges & Tree & Reg. & 10000 & 512 & - & 3 & 0 & 11.96 & 7.95 & 9.29e+05 & 11 \\
 R-33 & regression-num-medium-0-pol & Tree & Reg. & 10000 & 512 & - & 26 & 0 & 100 & 0 & 200 & 0 \\
 R-34 & regression-num-medium-0-superconduct & Tree & Reg. & 10000 & 512 & - & 79 & 0 & 185 & 0.00 & 22590 & 0 \\
 R-35 & regression-cat-large-0-diamonds & Tree & Reg. & 37758 & 512 & - & 6 & 3 & 9.84 & 5.79 & 79 & 0 \\
 R-36 & regression-cat-large-0-black\_friday & Tree & Reg. & 50000 & 512 & - & 4 & 5 & 10.08 & 5.23 & 20 & 0 \\
 R-37 & regression-cat-large-0-nyc-taxi-green-dec-2016 & Tree & Reg. & 50000 & 512 & - & 9 & 7 & 5.48 & 0 & 490 & 0 \\
 R-38 & regression-cat-large-0-particulate-matter-ukair-2017 & Tree & Reg. & 50000 & 512 & - & 3 & 3 & 5.57 & -1.64 & 260 & -4 \\
 R-39 & regression-cat-large-0-SGEMM\_GPU \_kernel\_performance & Tree & Reg. & 50000 & 512 & - & 3 & 6 & 8.09 & 2.66 & 3283.59 & 13.25 \\
 R-40 & regression-num-large-0-year & Tree & Reg. & 50000 & 512 & - & 90 & 0 & 2011 & 1922 & 47953.11 & -1.06e+04 \\

\bottomrule
\end{longtable}
}
{
\clearpage
\centering
\begin{longtable}[t]{p{1cm}|p{1cm}|p{1cm}|p{1cm}|p{1cm}|p{1cm}|cccccc}
\caption{Detailed results of all baseline models on all datasets.}
\label{tab:detail_res}
\\
\toprule
Dataset\newline{Index} & RAG+\newline{Phi3-GTL} & Phi3-GTL & RAG+\newline{KNN} & TabPFN\newline{-v2} & TabPFN\newline{-v1} & XGB. & CatBoost & LGB. & FTT. & MLP & TabR \\
\midrule
C-01 & $0.955$ & $0.776$ & $0.916$ & $0.968$ & $0.961$ & $0.954$ & $0.960$ & $0.936$ & $0.942$ & $0.959$ & $0.972$ \\
C-02 & $0.956$ & $0.894$ & $0.874$ & $0.954$ & $0.954$ & $0.952$ & $0.956$ & $0.960$ & $0.949$ & $0.959$ & $0.936$ \\
C-03 & $0.934$ & $0.751$ & $0.765$ & $0.941$ & $0.906$ & $\underline{0.945}$ & $0.944$ & $0.941$ & $0.936$ & $0.909$ & $\bm{0.949}$ \\
C-04 & $0.999$ & $0.968$ & $0.999$ & $\bm{1.000}$ & $\bm{1.000}$ & $0.998$ & $0.998$ & $0.998$ & $0.999$ & $\underline{1.000}$ & $0.999$ \\
C-05 & $0.939$ & $0.614$ & $0.874$ & $\bm{0.981}$ & $0.963$ & $0.969$ & $\underline{0.976}$ & $0.969$ & $0.963$ & $0.911$ & $0.966$ \\
C-06 & $0.771$ & $0.747$ & $0.795$ & $0.816$ & $0.811$ & $0.772$ & $0.796$ & $\bm{0.821}$ & $0.811$ & $0.807$ & $\underline{0.820}$ \\
C-07 & $\bm{1.000}$ & $0.996$ & $0.997$ & $0.995$ & $0.962$ & $1.000$ & $1.000$ & $0.999$ & $\underline{1.000}$ & $1.000$ & $1.000$ \\
C-08 & $0.990$ & $0.835$ & $0.986$ & $0.990$ & $0.989$ & $0.993$ & $0.992$ & $\underline{0.993}$ & $0.989$ & $0.990$ & $\bm{0.994}$ \\
C-09 & $0.999$ & $0.927$ & $0.998$ & $0.999$ & $0.998$ & $\bm{1.000}$ & $\underline{0.999}$ & $0.999$ & $0.998$ & $0.998$ & $0.999$ \\
C-10 & $0.815$ & $0.656$ & $0.821$ & $\bm{0.853}$ & $0.841$ & $0.835$ & $0.837$ & $\underline{0.842}$ & $0.835$ & $0.824$ & $0.833$ \\
C-11 & $0.892$ & $0.832$ & $0.887$ & $0.907$ & $0.895$ & $\underline{0.932}$ & $0.928$ & $\bm{0.933}$ & $0.903$ & $0.904$ & $0.911$ \\
C-12 & $0.936$ & $0.600$ & $0.915$ & $0.955$ & $0.937$ & $\underline{0.974}$ & $0.974$ & $0.973$ & $\bm{0.976}$ & $0.973$ & $0.964$ \\
C-13 & $0.736$ & $0.541$ & $0.755$ & $0.779$ & $0.743$ & $0.799$ & $0.796$ & $\underline{0.804}$ & $\bm{0.811}$ & $0.801$ & $0.798$ \\
C-14 & $0.981$ & $0.782$ & $0.962$ & $0.966$ & $0.947$ & $\underline{0.993}$ & $0.993$ & $0.993$ & $0.992$ & $0.991$ & $\bm{0.997}$ \\
C-15 & $\bm{0.906}$ & $0.751$ & $0.811$ & $0.899$ & $0.884$ & $0.895$ & $\underline{0.903}$ & $0.889$ & $0.863$ & $0.853$ & $0.887$ \\
C-16 & $0.943$ & $0.713$ & $0.851$ & $\underline{0.948}$ & $0.933$ & $0.930$ & $0.932$ & $0.928$ & $0.917$ & $0.914$ & $0.935$ \\
C-17 & $0.781$ & $0.545$ & $0.681$ & $\underline{0.921}$ & $0.731$ & $0.870$ & $0.882$ & $0.873$ & $0.809$ & $0.750$ & $\bm{0.934}$ \\
C-18 & $0.866$ & $0.562$ & $0.865$ & $\bm{0.917}$ & $0.868$ & $0.897$ & $0.893$ & $\underline{0.899}$ & $0.895$ & $0.874$ & $0.876$ \\
C-19 & $0.938$ & $0.852$ & $0.937$ & $\bm{0.950}$ & $0.939$ & $0.939$ & $\underline{0.945}$ & $0.938$ & $0.942$ & $0.938$ & $0.933$ \\
C-20 & $0.836$ & $0.672$ & $0.834$ & $0.860$ & $0.841$ & $0.856$ & $\underline{0.860}$ & $0.852$ & $0.845$ & $0.838$ & $\bm{0.860}$ \\
C-21 & $0.883$ & $0.674$ & $0.855$ & $\bm{0.930}$ & $0.921$ & $0.911$ & $0.917$ & $0.907$ & $0.914$ & $0.918$ & $\underline{0.922}$ \\
C-22 & $0.843$ & $0.587$ & $0.720$ & $0.787$ & $0.779$ & $\underline{0.885}$ & $0.845$ & $0.846$ & $0.835$ & $0.815$ & $\bm{0.937}$ \\
C-23 & $0.920$ & $0.714$ & $0.868$ & $0.944$ & $0.890$ & $\underline{0.962}$ & $0.956$ & $0.957$ & $0.927$ & $0.916$ & $\bm{0.984}$ \\
C-24 & $0.835$ & $0.754$ & $0.842$ & $\underline{0.850}$ & $0.847$ & $0.849$ & $\bm{0.852}$ & $0.848$ & $0.848$ & $0.845$ & $0.847$ \\
C-25 & $0.876$ & $0.638$ & $0.880$ & $0.877$ & $0.858$ & $0.891$ & $\bm{0.893}$ & $0.891$ & $0.886$ & $0.883$ & $\underline{0.892}$ \\
C-26 & $0.964$ & $0.572$ & $0.906$ & $0.956$ & $0.860$ & $0.985$ & $0.985$ & $0.986$ & $0.988$ & $0.984$ & $\bm{0.992}$ \\
C-27 & $0.851$ & $0.596$ & $0.821$ & $0.862$ & $0.832$ & $\underline{0.902}$ & $0.894$ & $0.899$ & $0.892$ & $0.888$ & $\bm{0.948}$ \\
C-28 & $0.803$ & $0.551$ & $0.807$ & $0.834$ & $0.776$ & $0.859$ & $0.854$ & $0.857$ & $\underline{0.861}$ & $0.859$ & $\bm{0.863}$ \\
C-29 & $0.973$ & $0.887$ & $0.968$ & $0.981$ & $0.979$ & $0.987$ & $0.987$ & $0.987$ & $\underline{0.987}$ & $0.987$ & $\bm{0.988}$ \\
R-01 & $\underline{0.071}$ & $\bm{0.070}$ & $0.088$ & $0.077$ & $-$ & $0.079$ & $0.083$ & $0.079$ & $0.093$ & $0.095$ & $0.090$ \\
R-02 & $0.015$ & $0.039$ & $0.034$ & $\bm{0.005}$ & $-$ & $0.006$ & $0.006$ & $0.007$ & $0.008$ & $0.008$ & $0.008$ \\
R-03 & $0.160$ & $0.190$ & $0.171$ & $\bm{0.142}$ & $-$ & $0.152$ & $0.156$ & $0.156$ & $0.152$ & $0.152$ & $0.150$ \\
R-04 & $0.017$ & $0.036$ & $0.041$ & $\underline{0.014}$ & $-$ & $0.015$ & $0.015$ & $0.016$ & $0.015$ & $0.015$ & $0.015$ \\
R-05 & $0.008$ & $0.049$ & $0.043$ & $\underline{0.007}$ & $-$ & $0.007$ & $0.007$ & $0.008$ & $0.008$ & $0.010$ & $0.008$ \\
R-06 & $0.094$ & $0.210$ & $0.088$ & $\bm{0.078}$ & $-$ & $0.081$ & $\underline{0.080}$ & $0.081$ & $0.080$ & $0.082$ & $0.082$ \\
R-07 & $0.137$ & $0.418$ & $0.191$ & $0.141$ & $-$ & $0.114$ & $0.117$ & $\underline{0.107}$ & $0.127$ & $0.125$ & $0.124$ \\
R-08 & $0.187$ & $0.358$ & $0.220$ & $\bm{0.126}$ & $-$ & $0.141$ & $0.134$ & $0.141$ & $0.150$ & $0.169$ & $\underline{0.127}$ \\
R-09 & $0.375$ & $0.647$ & $0.362$ & $\bm{0.283}$ & $-$ & $0.324$ & $0.316$ & $0.323$ & $0.313$ & $0.305$ & $\underline{0.289}$ \\
R-10 & $0.134$ & $0.292$ & $0.167$ & $\bm{0.091}$ & $-$ & $0.095$ & $0.097$ & $0.095$ & $0.098$ & $0.105$ & $0.094$ \\
R-11 & $0.701$ & $1.000$ & $0.704$ & $0.627$ & $-$ & $\bm{0.614}$ & $0.618$ & $\underline{0.617}$ & $0.654$ & $0.629$ & $0.630$ \\
R-12 & $0.846$ & $0.972$ & $0.848$ & $0.847$ & $-$ & $0.837$ & $0.838$ & $0.840$ & $\bm{0.825}$ & $0.846$ & $\underline{0.835}$ \\
R-13 & $0.111$ & $0.517$ & $0.109$ & $0.098$ & $-$ & $\bm{0.088}$ & $\underline{0.089}$ & $0.089$ & $0.090$ & $0.090$ & $0.090$ \\
R-14 & $\bm{0.005}$ & $0.014$ & $0.028$ & $0.008$ & $-$ & $0.009$ & $0.009$ & $0.012$ & $0.011$ & $0.013$ & $0.012$ \\
R-15 & $\underline{0.054}$ & $0.105$ & $0.057$ & $\bm{0.053}$ & $-$ & $0.055$ & $0.055$ & $0.056$ & $0.057$ & $0.058$ & $0.056$ \\
R-16 & $\bm{0.063}$ & $0.091$ & $0.096$ & $0.084$ & $-$ & $0.069$ & $\underline{0.068}$ & $0.072$ & $0.087$ & $0.084$ & $0.073$ \\
R-17 & $0.028$ & $0.061$ & $0.042$ & $0.019$ & $-$ & $0.021$ & $\bm{0.018}$ & $\underline{0.018}$ & $0.018$ & $0.021$ & $0.019$ \\
R-18 & $\bm{0.000}$ & $0.079$ & $0.031$ & $0.001$ & $-$ & $0.004$ & $0.001$ & $0.001$ & $0.005$ & $0.004$ & $0.004$ \\
R-19 & $0.021$ & $0.020$ & $0.021$ & $0.020$ & $-$ & $0.020$ & $\bm{0.020}$ & $0.020$ & $0.021$ & $\underline{0.020}$ & $0.020$ \\
R-20 & $0.129$ & $0.316$ & $0.169$ & $\underline{0.098}$ & $-$ & $0.120$ & $0.126$ & $0.131$ & $0.141$ & $0.140$ & $0.134$ \\
R-21 & $0.006$ & $0.029$ & $0.019$ & $\bm{0.000}$ & $-$ & $0.002$ & $0.001$ & $0.002$ & $0.003$ & $0.003$ & $0.002$ \\
R-22 & $0.175$ & $0.350$ & $0.160$ & $\underline{0.123}$ & $-$ & $0.127$ & $\bm{0.120}$ & $0.128$ & $0.134$ & $0.136$ & $0.135$ \\
R-23 & $0.011$ & $0.032$ & $0.014$ & $\bm{0.007}$ & $-$ & $0.008$ & $0.008$ & $0.008$ & $0.008$ & $0.008$ & $\underline{0.007}$ \\
R-24 & $0.087$ & $0.118$ & $0.085$ & $0.083$ & $-$ & $\underline{0.082}$ & $0.083$ & $\bm{0.082}$ & $0.083$ & $0.084$ & $0.084$ \\
R-25 & $0.375$ & $0.820$ & $0.514$ & $\bm{0.123}$ & $-$ & $0.137$ & $\underline{0.127}$ & $0.138$ & $0.128$ & $0.141$ & $0.133$ \\
R-26 & $0.016$ & $0.025$ & $0.015$ & $\bm{0.009}$ & $-$ & $0.010$ & $0.009$ & $\underline{0.009}$ & $0.010$ & $0.010$ & $0.009$ \\
R-27 & $0.111$ & $0.222$ & $0.138$ & $\bm{0.074}$ & $-$ & $0.082$ & $0.080$ & $0.082$ & $0.090$ & $0.099$ & $\underline{0.077}$ \\
R-28 & $0.139$ & $0.200$ & $0.139$ & $\bm{0.063}$ & $-$ & $0.071$ & $\underline{0.069}$ & $0.073$ & $0.145$ & $0.142$ & $0.144$ \\
R-29 & $0.076$ & $0.115$ & $0.077$ & $0.073$ & $-$ & $\underline{0.073}$ & $0.073$ & $0.073$ & $0.074$ & $0.075$ & $\bm{0.073}$ \\
R-30 & $0.039$ & $0.086$ & $0.038$ & $\bm{0.029}$ & $-$ & $0.031$ & $0.031$ & $\underline{0.031}$ & $0.033$ & $0.032$ & $0.032$ \\
R-31 & $0.016$ & $0.032$ & $0.023$ & $\bm{0.011}$ & $-$ & $0.012$ & $0.012$ & $0.012$ & $0.013$ & $0.014$ & $\underline{0.011}$ \\
R-32 & $0.005$ & $0.027$ & $0.007$ & $\bm{0.005}$ & $-$ & $0.006$ & $0.006$ & $0.006$ & $0.006$ & $0.006$ & $0.006$ \\
R-33 & $0.190$ & $0.556$ & $0.797$ & $\bm{0.034}$ & $-$ & $0.054$ & $0.059$ & $0.058$ & $\underline{0.042}$ & $0.078$ & $0.083$ \\
R-34 & $0.179$ & $0.719$ & $0.209$ & $0.164$ & $-$ & $\bm{0.162}$ & $\underline{0.162}$ & $0.162$ & $0.192$ & $0.177$ & $0.185$ \\
R-35 & $0.011$ & $0.029$ & $0.013$ & $\bm{0.007}$ & $-$ & $0.007$ & $0.007$ & $0.007$ & $0.007$ & $0.008$ & $0.008$ \\
R-36 & $0.030$ & $0.055$ & $0.031$ & $0.027$ & $-$ & $\underline{0.027}$ & $\bm{0.026}$ & $0.027$ & $0.027$ & $0.027$ & $0.027$ \\
R-37 & $\underline{0.200}$ & $0.435$ & $0.297$ & $0.240$ & $-$ & $0.230$ & $0.223$ & $0.223$ & $0.236$ & $0.244$ & $0.227$ \\
R-38 & $0.122$ & $0.153$ & $0.118$ & $0.112$ & $-$ & $0.109$ & $0.109$ & $\underline{0.108}$ & $0.110$ & $0.111$ & $\bm{0.107}$ \\
R-39 & $0.002$ & $0.010$ & $0.002$ & $\underline{0.002}$ & $-$ & $0.002$ & $0.002$ & $0.002$ & $0.002$ & $0.002$ & $0.002$ \\
R-40 & $0.004$ & $0.022$ & $0.004$ & $\bm{0.003}$ & $-$ & $0.003$ & $0.003$ & $0.003$ & $0.003$ & $\underline{0.003}$ & $0.003$ \\
\bottomrule
\end{longtable}
}

\end{document}